\documentclass[lettersize,journal]{IEEEtran}
\usepackage{amssymb}
 \usepackage{multirow}
\usepackage{bbm}
\usepackage{amsmath,amsfonts}
\usepackage{algorithmic}
\usepackage{algorithm}
\usepackage{array}
\usepackage[caption=false,font=normalsize,labelfont=sf,textfont=sf]{subfig}
\usepackage{textcomp}
\usepackage{stfloats}
\usepackage{url}
\usepackage{verbatim}
\usepackage{graphicx}
\usepackage{cite}
\begin{document}
\title{Knowledge Graph Reasoning with Self-supervised Reinforcement Learning}

\author{
    \IEEEauthorblockN{Ying Ma\IEEEauthorrefmark{1}, Owen Burns\IEEEauthorrefmark{1}, Mingqiu Wang, Gang Li, Nan Du, Laurent El Shafey, Liqiang Wang, Izhak Shafran, Hagen Soltau}
    \\ 

   \thanks{Ying Ma is with the School of Information and Electronics, Beijing Institute of Technology, Beijing 100081, China (e-mail: mayingbit2011@163.com). Ying is the corresponding author. \IEEEauthorblockA{\IEEEauthorrefmark{1}Denotes equal contribution.} \\
\hspace*{1em}Owen Burns and Liqiang Wang are with University
of Central Florida, Orlando, FL, USA. \\
\hspace*{1em} Mingqiu Wang, Laurent EI Shafey, Izhak Shafran and Hagen Soltau are with Google.
Nan Du is with the AIML division at Apple. Gang Li is with Orby AI. \\
© 2025 IEEE.  Personal use of this material is permitted.  Permission from IEEE must be obtained for all other uses, in any current or future media, including reprinting/republishing this material for advertising or promotional purposes, creating new collective works, for resale or redistribution to servers or lists, or reuse of any copyrighted component of this work in other works.
  }
}

\maketitle

\begin{abstract}
Reinforcement learning (RL) is an effective method of finding reasoning pathways in incomplete knowledge graphs (KGs). To overcome the challenges of a large action space, a self-supervised pre-training method is proposed to warm up the policy network before the RL training stage. To alleviate the distributional mismatch issue in general self-supervised RL (SSRL), in our supervised learning (SL) stage, the agent selects actions based on the policy network and learns from generated labels; this self-generation of labels is the intuition behind the name self-supervised. With this training framework, the information density of our SL objective is increased and the agent is prevented from getting stuck with the early rewarded paths. Our self-supervised RL (SSRL) method improves the performance of RL by pairing it with the wide coverage achieved by SL during pretraining, since the breadth of the SL objective makes it infeasible to train an agent with that alone. 
We show that our SSRL model meets or exceeds current state-of-the-art results on all Hits@k and mean reciprocal rank (MRR) metrics on four large benchmark KG datasets. 
This SSRL method can be used as a plug-in for any RL architecture for a KGR task. We adopt two RL architectures, i.e., MINERVA and MultiHopKG as our baseline RL models and experimentally show that our SSRL model consistently outperforms both baselines on all of these four KG reasoning tasks. Code for our method and scripts to run all experiments can be found in our GitHub at https://github.com/owenonline/Knowledge-Graph-Reasoning-with-Self-supervised-Reinforcement-Learning.

\end{abstract}
\section{Introduction}
Knowledge graphs (KG) support a variety of downstream tasks, such as question answering and recommendation systems \cite{yu-etal-2022-kg,he-etal-2017-learning,moon-etal-2019-opendialkg,huang2019knowledge}. The objective of \textit{knowledge graph completion} (KGC) (also known as \textit{knowledge graph reasoning}, KGR)  is to infer missing information for a given KG. KGC is a pertinent problem, as practical KGs often fail to include all relevant facts. The two main approaches to KGC are embedding-based methods and path-based methods, the latter having the advantage of providing interpretable reasoning paths on the graph. Across path-based methods, previous work shows that path-ranking-based methods and deep reinforcement learning (RL) based methods \cite{minerva,shen2018m,LinRX2018:MultiHopKG} achieve state-of-the-art results. Path-ranking-based methods are usually formulated for the link prediction task (predicting the existence of a path between two entities) and most RL-based methods are usually formulated for the query answering task (predicting a logical end entity given a start entity and a relation), although some RL-based methods such as \cite{wenhan_emnlp2017} can also be adapted to both. In this paper, we focus on the query answering task to show the effectiveness of our proposed method.

Unlike the common applications of RL such as robotics and video games, formulating KGC as an RL problem presents a challenge of large action spaces. For example, many nodes in the current KG datasets connect to hundreds or thousands of nodes. Similar cardinalities often arise in real world scenarios as well, such as constructing a knowledge graph to represent a large medical corpus. In this case, entities could include diseases, symptoms, treatments, and medications, while relations might define interactions such as 'causes', 'alleviates', or 'contraindicated with'. Accurately inferring missing links, such as identifying an effective treatment for a rare combination of symptoms, relies heavily on the ability to make unintuitive but correct connections, which heavily relies on efficient exploration of the full breadth of the graph. To improve exploration efficiency, we propose a self-supervised RL approach (SSRL), where the RL network is warmed up in a supervised manner in order to give the agent a lead for queries with large action space where RL alone performs poorly in.

Combining SL with RL is not new in the RL research community (in the RL research community, it is named imitation learning, learning from demonstration, or more specifically, behavior cloning \cite{ravichandar2020recent}). However, there is a notorious issue of distributional mismatch between generated paths and real policy in SSRL (in general SSRL, instead of exploring unseen states, the agent may keep on exploiting the visited states found by the generated paths). KGs represent a static and fully deterministic environment in which answer entities typically lie in the near neighborhood of the start entity. For example, in the FB15K-237 dataset \cite{toutanova-etal-2015-representing}, 99.8\% queries have correct answers within the 3-hop neighborhood. Such a structure enables finding ground-truth paths and generating labels for most nodes. Based on this observation, in order to prevent the agent from getting stuck with the early rewarded paths, we grant our agent more freedom of exploration and increase the information density of our SL objective by two modifications: in the SL stage, 1) instead of following the randomly generated paths, our agent selects actions based on probability defined by the policy network; 2) our agent can be trained by minimizing the distance between the policy network output and labels instead of maximizing the expected reward since we have labels for each state (we collect all correct paths and generate a state-label pair instead of a state-action pair for each state). With these two modifications, our agent does not need to always follow a specific generated path and can get more contextual information, hence, the distributional mismatch issue is solved to some extent. In summary, our contributions can be summarized as follows.

1) We propose an SSRL framework for the query answering task. This SSRL method can be used as a plug-in for any RL architecture for the KGC task. We experimentally show that our SSRL method achieves state-of-the-art results on four large benchmark KG datasets and the SSRL consistently outperforms the baseline RL architecture to which it is applied on four large benchmark KG datasets. 

2) To solve the distributional mismatch problem in the general SSRL framework, we analyze the pros and cons of SL and RL in terms of
coverage, learning speed, and feasibility in static and deterministic KG environments. Since creating state-label pairs is feasible for KG environments (in common RL environments such as video games, only some positive trajectories can be generated), 
we use a different exploration strategy and training loss function from the general SSRL framework to solve the distributional mismatch problem to some extent. 

4) We compare our proposed SSRL paradigm with the general SSRL used in DeepPath \cite{wenhan_emnlp2017} on the link prediction problem and show the superiority of our SL strategy experimentally.

\section{Related Work}
KGC task can be solved with embedding-based methods \cite{cao2022geometry}\cite{chao2020pairre}\cite{li2022does}\cite{song2021rot}\cite{DBLP:journals/corr/YangYHGD14a},\cite{dettmers2018convolutional}, \cite{toutanova-etal-2016-compositional} and path-based methods. Compared to path-based methods, embedding-based methods cannot capture more complex reasoning patterns and are less interpretable. Across path-based methods, path-ranking-based methods and deep reinforcement learning (RL)-based methods achieve state-of-the-art results. Path-ranking-based methods formulate KGC as a link prediction task. Path-ranking-based methods gather paths by performing random walks \cite{das-etal-2017-chains}, \cite{gu2015traversing}, \cite{lao-etal-2011-random}, \cite{neelakantan-etal-2015-compositional}, \cite{toutanova-etal-2016-compositional} or using attention techniques \cite{liu2020path}\cite{yang2017differentiable}\cite{rocktaschel2017end}. These methods are computationally expensive, as the entire graph needs to be searched or accessed. 
Recently, several RL-based methods have been proposed to formulate KGC as a query answering task. MINERVA \cite{minerva} uses a policy gradient to explore paths and CPL \cite{fu-etal-2019-collaborative} extends to large KG datasets such as FB60K. Observing the MINERVA sparse reward problem, M-Walk \cite{shen2018m} proposes off-policy learning in which search targets are scored based on a value function based on search history. MultiHopKG \cite{LinRX2018:MultiHopKG} uses embedding-based methods for reward shaping. In comparison, our pre-training with seeding labels has better coverage, therefore achieving higher prediction accuracy. Combining SL with RL is named learning from demonstration in the RL research community \cite{florence2022implicit}\cite{osa2018algorithmic}\cite{ho2016generative}. DeepPath \cite{wenhan_emnlp2017}  is the first work to adopt this general SSRL training paradigm for both link prediction and query answering task.

\section{Problem Formulation}
In this section, we first define the query answering task on KG, then we describe how to formulate it as a RL problem.
\subsection{Problem definition}
A knowledge graph is represented as $\mathcal{G}=(\mathcal{E},\mathcal{R})$, where $\mathcal{E}$ is the set of entities $e \in \mathcal{E}$ and $\mathcal{R}$ is the set of directed relations $r \in \mathcal{R}$ between two entities. The query answering task can be defined as follows. Given a query $(e_s, r_q, e_q)$, where $e_s$ is the source entity, $r_q$ is the relation of interest, and $e_q$ is the target entity which is unknown to the agent, the objective of query answering is to infer $e_q$ by finding paths starting from $e_s$, i.e., $\{(e_s,r_0,e_1), (e_1,r_1,e_2)$ ,..., $(e_t,r_t,e_{t+1}), ...(e_n,r_n,e_q)\}$. For example, if $e_s$ is "Tom Brady" and $r_q$ is "coached by", the agent might take path {(Tom Brady, played for, Buccaneers), (Buccaneers, head coach, Todd Bowles), (Todd Bowles, no-op, Todd Bowles)}, inferring that $e_q$ is Todd Bowles. 

\subsection{Reinforcement learning formulation}
The path finding process can be viewed as a partially observed deterministic Markov Decision Process (MDP) where the agent starts from the source entity $e_s$ and sequentially selects the outgoing edges for predefined $T$ steps. MDP is formally defined by the 5-tuple $(\mathcal{S}, \mathcal{O}, \mathcal{A}, \mathcal{R}, \delta)$, where $\mathcal{S}$ is the set of states, $\mathcal{O}$ is the set of observations, $\mathcal{A}$ is the set of actions, $\mathcal{R}$ is the reward function, and $\delta$ is the state transition function. We elaborate the specific meaning of each of those as follows.
\paragraph{States and Observations}
The state $s_t \in \mathcal{S}$ at time $t$ is a 4-tuple $s_t = (e_s,r_q,e_q,e_t)$, where $e_t$ is the entity visited at time $t$. The environment is only partially observed, as the agent does not know the target entity $e_q$. So, the observation $o_t \in \mathcal{O}$ at time $t$ is a 3-tuple $o_t =(e_s,r_q,e_t)$, where $e_s$ and $r_q$ are the context information for each query and $e_t$ is the state-dependent information.
\paragraph{Actions}
The possible actions $A_t \in \mathcal{A}$ at time $t$ are all the outgoing edges connected to $e_t$, $A_t=\{(r,e)|(e_t,r,e) \in \mathcal{G} \}$. A self-loop action NO\_OP is added to every $A_t$ to give the agent the option to stay on a specific node for any number of steps in case the correct answer is reached at $t<T$. And a reverse link is added to each triple, that is, adding $(e_1,r^{-1},e_2)$ to $(e_2,r,e_1)$, to allow our agent to undo a potentially wrong decision.

\paragraph{Rewards}
The agent receives a terminal reward 1 if it reaches the correct answer $e_q$ at time $T$ and 0 otherwise, i.e., 
$R(s_T)=\mathbbm{1}\{e_T=e_q\}$.
\paragraph{Transition}
The environment is completely determined by the graph, so it evolves deterministically. The transition function is defined by $\delta (s_t,a_t)=s_{t+1}$, where $a_t$ is the action selected at time $t$.

\section{Approach}
In this section, we first describe the policy network architecture. Then the label generation process is described for the SL training stage. Finally, our two-stage training framework is explained.

\begin{figure*} 
    \centering
    \includegraphics[scale=0.53]{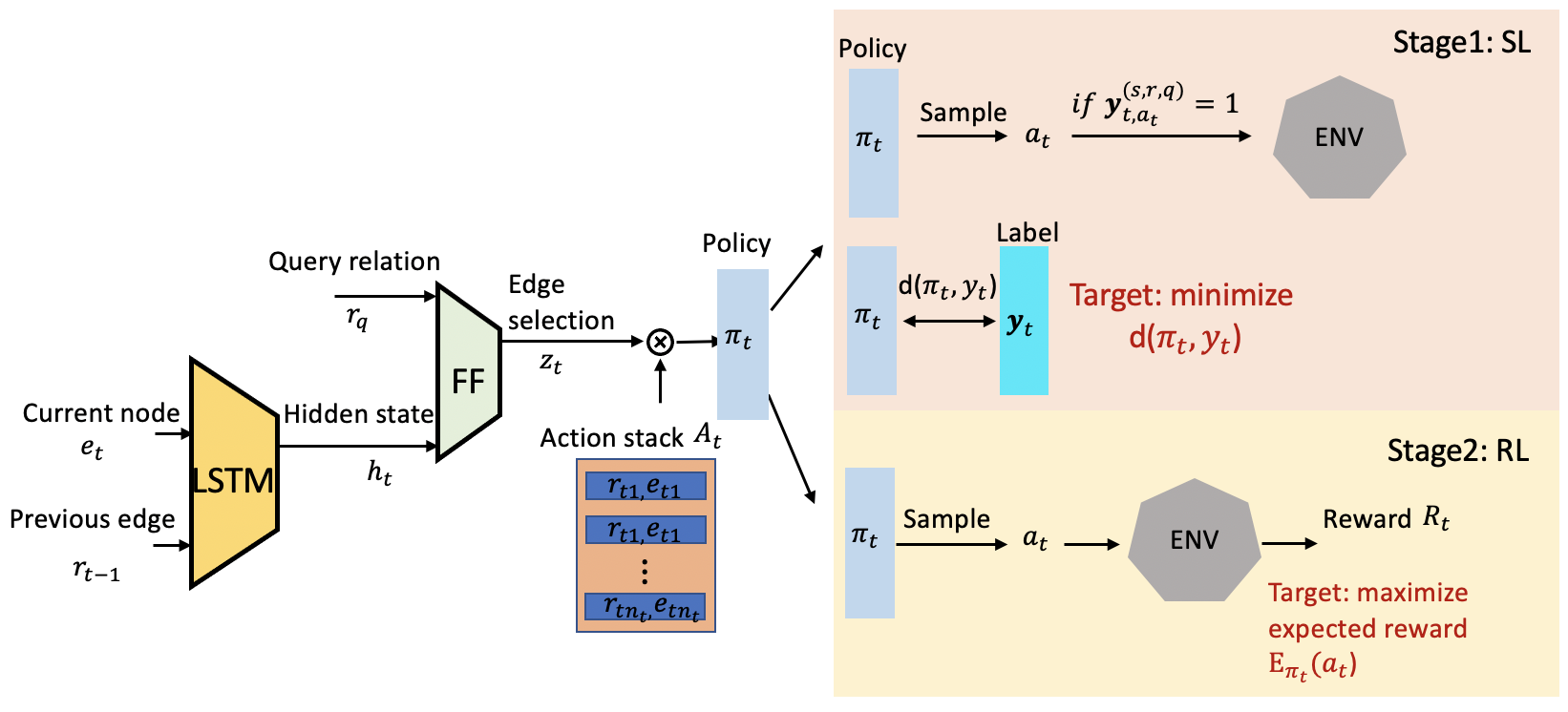}
    \caption{System architecture of SSRL}
    \label{fig:sys}
\end{figure*}

\subsection{Policy network}
\label{sec:policynetwork}

A neural network is used to predict the policy $\pi_t = \mathcal{P}(\mathcal{S}_{t},\mathcal{A}_{t})$, i.e., the probability of possible actions at time $t$. The network architecture is shown in Fig.~\ref{fig:sys}.

An LSTM \cite{hochreiter1997long} is adopted to encode state-dependent information in a vector $\mathbf{h_t}$.
\begin{equation}
    \mathbf{h_t} = \mathrm{LSTM}([\mathbf{r_{t-1}};\mathbf{e_{t}}]),
\end{equation}
where $\mathbf{e_{t}}\in \mathbbm{R}^d$ and $\mathbf{r_{t}}\in \mathbbm{R}^d$ are vector representations of the entity and the selected edge at time $t$. And $[;]$ denotes the vector concatenation.

Then, the history embedding $\mathbf{h_t}$ and the query relation embedding $\mathbf{r}_q$ are concatenated before being fed into a feedforward network with the ReLU non-linearity parameterized by $\mathbf{W_1}$. Since $\mathbf{r}_q$ is context information and time-invariant, a feedforward network is capable of encoding and extracting information from it. The action space is encoded by stacking the embeddings of all possible actions into a matrix $\mathbf{A_t}\in \mathbbm{R}^{n_t \times 2d}$. Here $n_t$ is the number of actions of the entity $e_t$, which varies for different entities $e_t$. The policy is calculated as follows,
\begin{align}
    \mathbf{z}_t= \mathbf{W_2}\mathrm{ReLU}(\mathbf{W_1}[\mathbf{h_t};\mathbf{r_q}])\\
    \pi_t=\sigma (\mathbf{A_t}\times \mathbf{z}_t).
\end{align}
Here, $\sigma$ is the softmax function and $\mathbf{z}_t$ is the action selection vector. The probability of taking action $a_t$ is determined by the similarity (inner product) between $\mathbf{z}_t$ and the corresponding element in $\mathbf{A_t}$. To guarantee the stability of training, the action space of $\mathbf{A_t}$ is truncated by a constant.

\subsection{Label generation}
\label{seeding}
\begin{figure*} 
    \centering
    \hspace{-25mm}\includegraphics[scale=0.45]{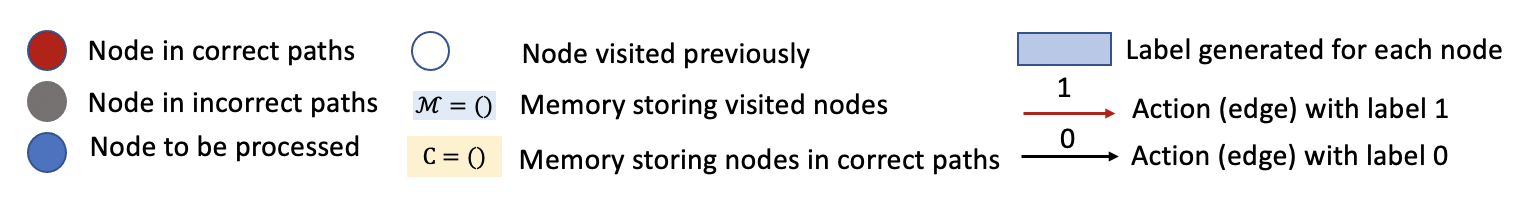}\\
    \includegraphics[scale=0.27]{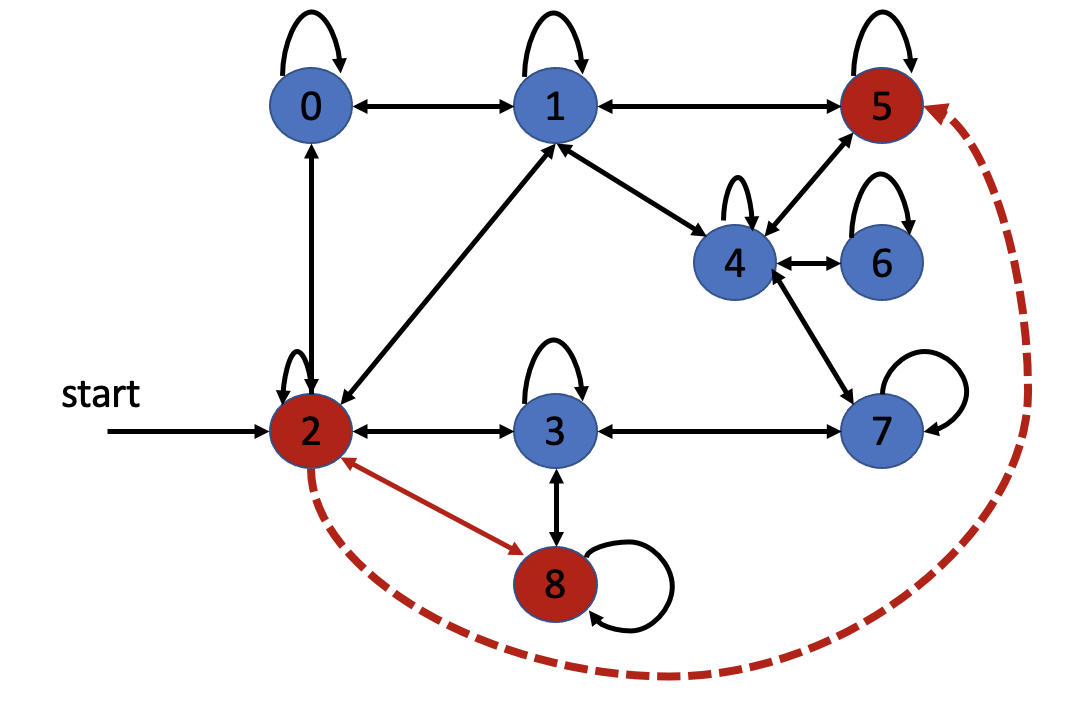}
    \includegraphics[scale=0.27]{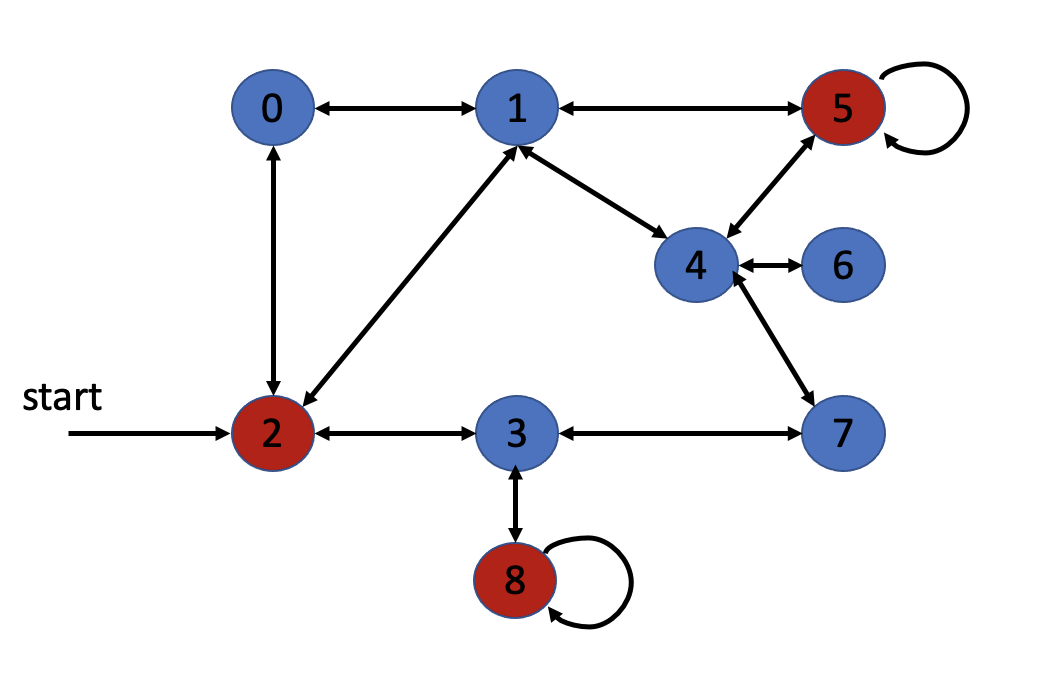}
    \includegraphics[scale=0.27]{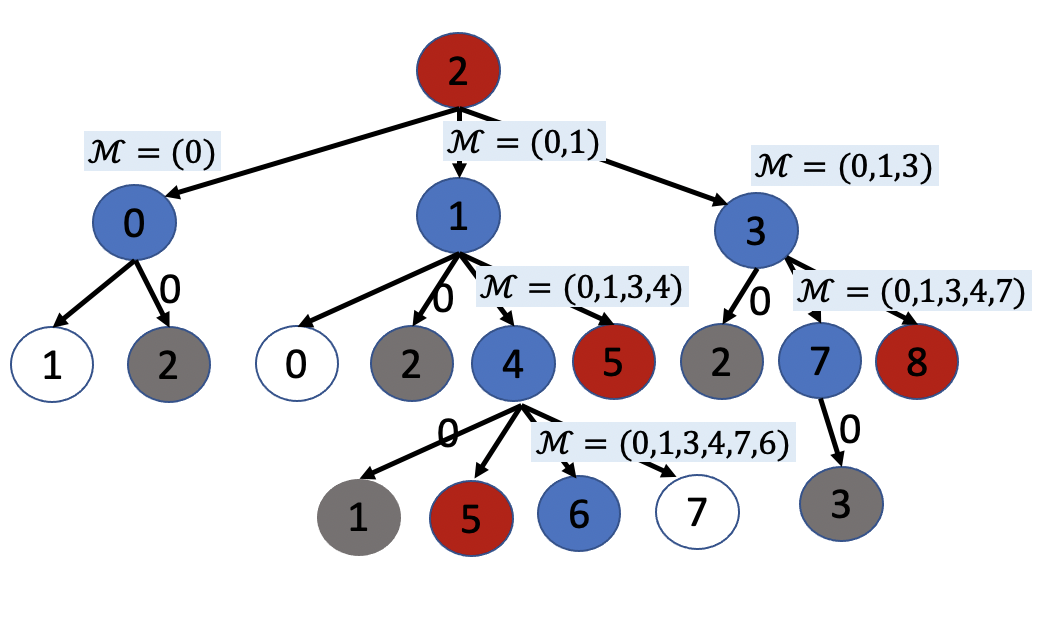}\\
    (a)\hspace{45mm} (b) \hspace{45mm}(c)\\
    \includegraphics[scale=0.27]{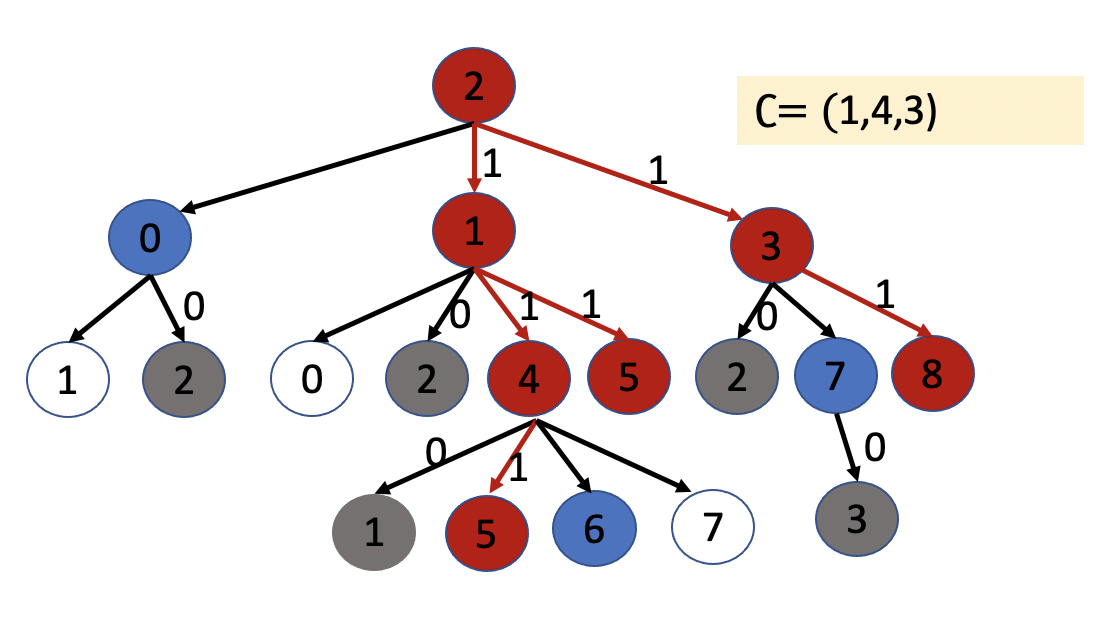}
    \includegraphics[scale=0.27]{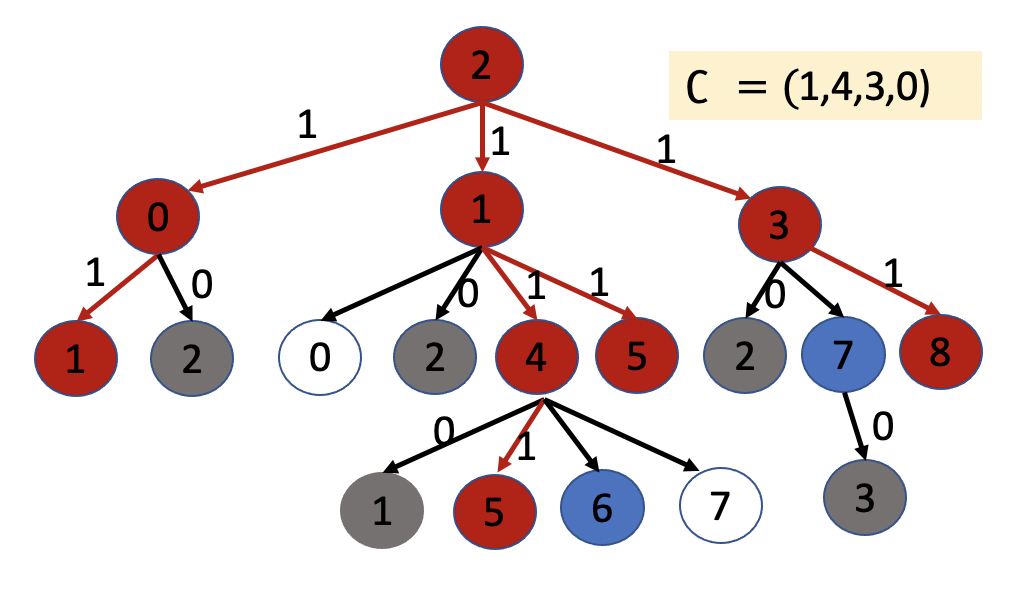}
    \includegraphics[scale=0.37]{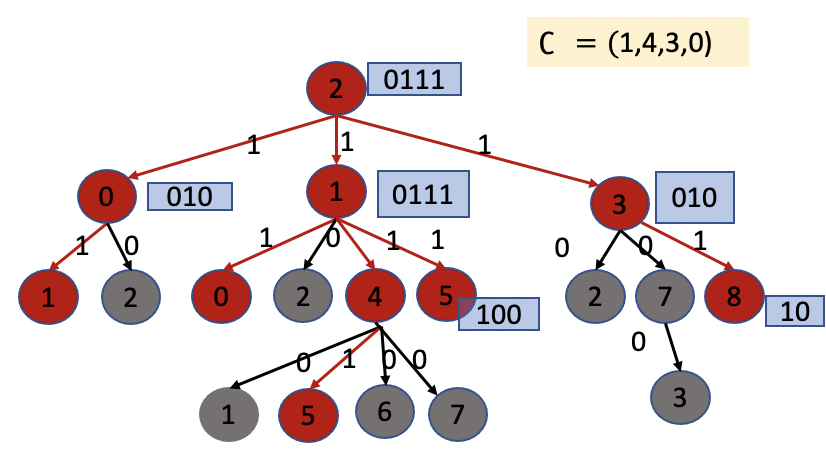}\\
    (d)\hspace{45mm} (e) \hspace{45mm}(f)\\
    \caption{seeding label generation process. 
    (a) 3-hop neighborhood of the start entity $e2$; the red dashed line represents the missing link that must be inferred.
    (b) Step1: remove the link between $e_2$ and $e_5$ and self-loop of all nodes except for the ones in $E_{\mathrm{all}}$;
    (c) First round of traverse. The nodes visited before are added to the set $\mathcal{M}$.
    (d) Marking nodes on correct paths as red.
    (e) and (f) Generating labels for left nodes in $\mathcal{C}$.
    }
    \label{fig:my_label}
\end{figure*}
In this section, we explain the label generation process for each node. In KGs, the same start entity $e_s$ and relation $r$ may connect to many different target entities, i.e., one to many connections. For example, in the freebase KG, the target entity of (CEO, job position in organization, ?) could be Merck \& Co., Google LLC., etc. Here, we use $E_{\mathrm{all}}$ to represent the set of all end entities of $(e_s,r,?)$. To warm up the policy network to wider coverage, for query $(e_s,r,e_q)$, the paths connecting $e_s$ with all entities in $E_{\mathrm{all}}$ are marked as correct paths. The labels for all nodes on the correct paths are generated with breadth-first search (BFS). 

The generated labels of entity $e_t$ for query $(e_s,r,e_q)$ is represented in form, 
\begin{equation}
    L((s,r,q),e_t)= \mathbf{y}_t^{(s,r,q)}\in \mathbbm{R}^{n_t}.
\end{equation}
Here, we use $\mathbf{y}_t^{(s,r,q)}$ to represent the label of the entity visited at the time step $t$ and $\mathbf{y}\mathrm{I}^{(s,r,q)}$ to represent the label of the entity $e\mathrm{I}$. The superscript $(s,r,q)$ applied is to show the labels is for query $(e_s,r,e_q)$. The first element of $\mathbf{y}_t^{(s,r,q)}$ represents self-connection. So, the label of entity $e_t\in E_{all}$ is $\mathbf{y}_t^{(s,r,q)}=[1,0,0,...]^\mathrm{T}$ (i.e., the first element is one and all other elements are zeros, the length of the vector is the same as the number of possible actions at time $t$), indicating that this is the correct answer. 

All other nodes lying on the correct paths are found and labeled with breadth first search (BFS) with a memory of the nodes visited $\mathcal{M}$ to avoid repeatedly generating labels for the same node. We use an example of generating labels for the query $(e2,r1,e5)$ to explain the method.  
Fig.~\ref{fig:my_label}(a) is a 3-hop neighborhood of the start entity $e_2$. Since there is a fact $(e2,r1,e8)$ existing on the graph ($E_{all}=(e5,e8)$), paths connecting $e2$ with both $e5$ and $e8$ are all marked as correct paths.

\textit{Step1:} The first step is to remove the self-loop of all nodes except for nodes in $E_{\mathrm{all}}$, i.e., $e5$ and $e8$, as shown in Fig.~\ref{fig:my_label}(b). The self-loop for nodes in $E_{\mathrm{all}}$ is kept to allow the agent to remain at these target nodes once it reaches. The labels for $e5$ and $e8$ are $\mathbf{y}5^{(2,1,5)}=[1,0,0]$ and $\mathbf{y}8^{(2,1,5)}=[1,0]$ respectively. The first element of the labels represents self-connection.  

\textit{Step2:} BFS is implemented to find entities in $E_{all}$. To avoid going back, the labels for the parent nodes are marked directly as zero without further searching, such as $e2$ for the nodes $e0$, $e1$, $e3$ shown in gray in Fig.~\ref{fig:my_label}(c). Visited nodes are also added to the set $\mathcal{M}$ to avoid being processed more than once. For example, since $e0$ and $e1$ have already been visited and added to $\mathcal{M}$ in the first hop, they are marked with empty circles indicating no further search. Finally, three paths are found: $e2\rightarrow e1\rightarrow e4 \rightarrow e5$, $e2 \rightarrow e1 \rightarrow e5$, $e2\rightarrow e3\rightarrow e8$ as shown in Fig.~\ref{fig:my_label}(d). And all the nodes in the correct paths are stored in the set $\mathcal{C}=(e1,e4,e3)$. 

\textit{Step3:}  For each node $e\mathrm{I} \in \mathcal{C}$, add all its parent nodes to $\mathcal{C}$ and repeat this process recursively until the source node is reached. For example, for node $e1$, since there is a path $e2\rightarrow e0\rightarrow e1$, node $e0$ is added to $\mathcal{C}$ as shown in Fig.~\ref{fig:my_label}(e).

\textit{Step4:} In the last step, the labels for nodes $e\mathrm{I} \in \mathcal{C}$ are generated. It should be noted that since the agent only selects the edges in the correct paths in the SL stage, as explained in the next section, there is no need to generate labels for nodes $e\mathrm{I} \notin \mathcal{C}$ such as $e6$ and $e7$. All unlabeled edges connecting to nodes in $\mathcal{C}$ are labeled as ones. All other edges are labeled zeros. And the labels for all nodes in $\mathcal{C}$ are shown in Fig.~\ref{fig:my_label}(f). The first element of the nodes $e_0,e_1,e_2,e_3$ is set as 0 to encourage the agent to explore other nodes in the environment, as the target nodes have not been reached yet.

\subsection{Training framework}
\label{train framework}
Our framework consists of an SL stage followed by an RL stage, enabling the agent to gain familiarity with the full breadth of the knowledge graph before learning the paths which maximize reward. The agent trains using the SL stage objective for a pre-set number of steps before the model weights are saved and training continues with the RL stage objective for the remaining steps.

\textbf{SL stage} The policy network is trained in a SL manner using the labels generated with breadth first search (BFS). The label generation process is explained in Section \ref{seeding}. Here, we use $\mathbf{y}_t$ to represent the label of the entity visited at the time step $t$. The elements in $\mathbf{y}_t$ are those if the related outgoing edges connect to end entities within several hops. It should be noted that our label generation process is different from that in DeepPath \cite{wenhan_emnlp2017}, where an individual random correct path between $e_s$ and $e_q$ is generated and used separately. 

The agent samples the action based on the policy $\pi_t$ approximated by the network.
\begin{equation}
    a_t \sim \mathrm{Categorical}(\pi_t),
\end{equation}
In order to travel only on the correct paths, the agent applies the selected action to the environment only if the label of that action is one, i.e., $\mathbf{y}_{t,a_t}=1$. Otherwise, the agent stays on the current node. The goal is to minimize the distance between $\pi_t$ and $\mathbf{y}_t$. Here we use the cross entropy loss as the cost function,
\begin{align}
L_{SL}(\theta) =&d(\pi_t,\mathbf{y}_{t})=-\frac{1}{n_t}\sum_{i=1}^{n_t}(\mathbf{y}_{t,i}\mathrm{log}\pi_{t,i}\nonumber\\
    &+(1-\mathbf{y}_{t,i})\mathrm{log}(1-\pi_{t,i})),
\end{align}
here $i$ is the index of the element in vector $\mathbf{y}_{t }^{(s,r,q)}$ and $\pi_{t}$. And the parameters $\mathbf{\theta}$ of the policy network are updated with stochastic gradient descent.
For example, if the label for entity $e_t$ is $\mathbf{y}_t=[0,1,0,1]^T$. If the agent samples the action $a_t=3$ based on the policy network, it traverses to the second node connected to $e_t$; if the sampled action is $a_t=2$, then instead of traversing to the third node, it just stays at the current entity since $a_t=3$ is not a valid action.

Note that the environment exploration strategy and loss function used in the SL training stage here are different from those in the general SS-RL framework. To be specific, instead of following the generated paths, our agent selects actions based on the probability defined by the
policy network. In this way, we grant our agent more
freedom of exploration and prevent it from getting
stuck with the early rewarded paths. Since collecting all correct paths and creating state-label pair is feasible for a deterministic environment like the KGs, our network is trained by minimizing
the distance between the policy network output and labels
instead of maximizing the expected reward. The information density
of our SL objective is higher than the expected reward used in the general SS-RL framework as our agent can get more
contextual information. Therefore, the distributional mismatch
issue is solved to some extent. The superiority of our SL training strategy is shown experimentally in section \ref{deeppath comp} by comparing with Deeppath, which implements the general SS-RL on link prediction tasks on KGs.

\textbf{RL stage} The action at time $t$ is sampled from the policy distribution and applied to the environment. 

The reward $R_t$ is collected at time $t$. The goal is to maximize the reward expectation, $L_{RL}(\theta)=\mathbbm{E}_{\pi}(R_t)$.
With the likelihood ratio trick, we write the derivative of the expectation of reward as the expectation of the derivative of reward and take one sample to update the parameters $\mathbf{\theta}$ of the policy network with stochastic gradient ascent,
\begin{equation}
    \theta \leftarrow  \theta     + \eta \triangledown_{\theta} L_{RL}(\theta),
\end{equation}
here $L_{RL}(\theta)$ is the loss for each sample at time $t$,
\begin{equation}
    L_{RL}(\theta)= \mathrm{log}\pi_{t}(a_t)G_t,
\end{equation}
\begin{equation}
    G_t=\sum_{k=t}^{T}\gamma ^{k-t}R_k,
\end{equation}
where $G_t$ is the discount accumulate reward with discount factor $\gamma$.

\textbf{SL v.s. RL in a deterministic environment}
To prove the superiority of combining SL and RL for the KGC task, we analyze the pros and cons of SL and RL in terms of coverage, learning speed, and feasibility.

\textit{Coverage}~~Since all the correct actions are marked as ones and the loss function of SL is to minimize the distance between the policy and the label at each decision step, the SL agent aims to find all possible paths from the start entity to the target entity. However, the goal of the RL agent is to find at least one correct path, so it can be rewarded when it reaches the target entity. There is no motivation for the RL to find more correct paths as long as it already finds one. The cross-entropy loss of the SL agent allows it to learn as if it were taking every action possible at each step of its training while the RL agent is constrained to taking and learning from only one action. Hence, the SL agent tries to cover more correct paths than the RL agent. This wide coverage allows SL agents to perform well in situations that RL agents perform poorly in, primarily unfamiliar situations with large numbers of possible actions.

\textit{Learning Speed}~~
While SL agents in general learn faster than RL ones, in our implementation the RL agent learns faster since the SL and RL portions of training have different objectives, i.e. RL aims to find at least one correct path while SL aims to minimize the difference between the policy outputted by the agent and a correct answer label of all valid actions from the current state. In both cases, actions are selected based on the policy estimated by the deep neural network.

\textit{Feasibility} Another disadvantage of SL is its label-generating process, which consumes additional computational resources. For large KGs such as FB15K-237 and FB60K\footnote{\url{https://github.com/thunlp/JointNRE}}, generating labels for all training queries borders on infeasible, making pure SL impractical. RL methods are conductible for most KGs as there is no requirement to generate and store path labels.

In order to combine the strengths of SL and RL, we propose this 2-stage SSRL training paradigm. Compared to the pure-RL method, our SSRL agent is warmed up to have a wider coverage so it achieves better performance. We also need to point out that although this SSRL framework is feasible for current KG datasets, the disadvantage of our SSRL is its extra label generation and SL training process.

\section{Experiment}

\begin{table*}[ht]
\centering
\caption{Statistics of datasets used in experiments
}
\begin{tabular}{lrrrrrrrrr}
\hline
\multirow{2}{4em}{\textbf{Dataset}} & \multirow{2}{4em}{\textbf\# Ent} & \multirow{2}{4em}{\textbf\# Rel}& \multirow{2}{4em}{\textbf\# Fact}& 

\multicolumn{2}{c}{Degree}\\
  & & & & avg.& median\\
\hline
FB15K-237 & 14,505 & 237 & 272,115  & 19.74 & 14 \\
WN18RR & 40,945 & 11 & 86,835  & 2.19 &2\\
NELL-995 &  75,492 & 200 & 154,213 & 4.07 &1 \\
FB60K &  69,514 & 1,327& 268,280  & 4.35 &4 \\
\hline
\end{tabular}

\label{dataset}
\end{table*}

\subsection{KG datasets statistics }

\label{KG statistic}
In order to analyze the performance of our proposed SSRL on different types of graphs, graph statistics such as the number of entities, edges and facts in each KG and the average and median graph degrees are summarized in Table~\ref{dataset}. Among all these graphs, FB15K-237 has the least number of entities, but the highest degree. It is more densely connected than other KGs, so it is the most challenging dataset for our SSRL method as generating labels for it is both time and resource consuming. 

\subsection{Model comparison}

\begin{table}[]
\centering
\caption{Hyperparamters}
\begin{tabular}{lllll}
\hline
            & \multicolumn{2}{l}{$\beta$} & \multicolumn{2}{l}{$\lambda$} \\
            & SL           & RL        & SL           & RL          \\
\hline
NELL-995    & 0.02         & 0.05      & 0.02         & 0.02        \\
FB15K-237   & 0.0002       & 0.02      & 0.02         & 0.02        \\
WN18RR      & 0.02         & 0.05      & .002         & 0.05        \\
FB60K-NYT10 & 0.02         & 0.2       & 0.02         & 0.02       \\
\hline
\end{tabular}

\label{hyperparameter}
\end{table}

\begin{table*}
\centering
\caption{Comparison of our SSRL method with state-of-art path based query answer methods. Our SSRL pre-training is applied to MINERVA and MultihopKG. All the metrics here are multiplied by 100. 
}
\begin{tabular}{llcccccc}
\hline
\multirow{2}{4em}{\textbf{Data}} & \multirow{2}{4em}{Metric}& \multirow{2}{4em}{NeuralLP} &\multirow{2}{4em}{M-Walk}& \multicolumn{2}{c}{MINERVA}  &\multicolumn{2}{c}{MultiHopKG} \\
&&&&Original&Ours&Original&Ours\\
\hline
\multirow{4}{4em}{FB15K-237} & HITS@1&16.6& 16.5& 21.7& 22.3(+0.6)&30.8&\textbf{31.9} (+1.1)\\
& HITS@3 & 24.8&24.3&32.5&34.5(+2)&43.3&\textbf{44.9}(+1.6)\\
& HITS@10 &34.8&-&44.5&47.6(+3.1)&55.6&\textbf{56.8}(+1.2)\\
& MRR &22.7&23.2&29.3&30.5(+1.2)&39.1&\textbf{40.4}(+1.3)\\
\hline
\multirow{4}{4em}{WN18RR} & HITS@1&37.6&41.4&44.9&\textbf{45.9}(+1)&39.0&38.2(-0.8)\\
& HITS@3 &46.8&44.5&49.3&\textbf{50.0}(+0.7)&44.8&45.9(+1.1)\\
& HITS@10 &\textbf{65.7}&-&54.6&55.4(+0.8)&50.5&51.1(+0.6)\\
& MRR &46.3&43.7&48.0&\textbf{49.1}(+1.1)&43.1&43(-0.1)\\
\hline
\multirow{4}{4em}{NELL-995} & HITS@1&-&68.4&65.3&\textbf{71.4} (+6.1)&65.7&69.4(+3.7)\\
& HITS@3 &-&81.0&79.7&80.5(+0.8)&84.9&\textbf{87.5} (+2.6)\\
& HITS@10 &-&- &82.3 &83.5(+1.2)&88.0&\textbf{89.1}(+1.1)\\
& MRR &-&75.4 &72.6&76.2(+3.6)&75.8&\textbf{78.9}(+3.4)\\
\hline
\end{tabular}

\label{performance comp}
\end{table*}

\begin{table}
\caption{Comparison of MINERVA and our SS-MINERVA(Semi-supervised MINERVA) on FB60K.}
\centering
\begin{tabular}{lcc}
\hline
 Metric&  MINERVA & Ours(SS-MINERVA) \\
 \hline
 HITS@1&36.2&\textbf{37.0}(+0.8)\\
 HITS@3 &42.0&\textbf{43.6}(+1.4)\\
 HITS@10 &48.4&\textbf{50.1}(+1.7)\\
 MRR &40.2&\textbf{41.4}(+1.2)\\
\hline
\end{tabular}

\label{performance compfb60k}
\end{table}

We evaluated our SSRL method on four large KGs, i.e., FB15K-237, WN18RR \cite{NIPS2013_1cecc7a7}, NELL995 \cite{wenhan_emnlp2017}, and FB60K, with different properties and from different domains. We compare our SSRL with other path-based methods (NeuralLP, M-Walk, MINERVA, and MultiHopKG). Since MINERVA and MultihopKG\footnote{We use the ConvE version of the MultihopKG.} achieve the state-of-the-art except Hits@10 on WN18RR among all path-based methods, we adopt these two as our baselines and apply our SSRL on top of them, respectively (we only use MINERVA as our baseline on FB60K as MultihopKG does not scale up on dataset with large relation numbers). 

\subsubsection{Hyperparameter setting}
For all experiments we kept the network parameters (e.g. number and size of layers, initialization method which we kept as Xavier, etc.) the same as the papers from which they were sourced.

\paragraph{MINERVA baseline}
Tests were conducted on an Nvidia Tesla V100 GPU and took ~18 hours for NELL-995 and WN18RR, ~24 hours for FB60K, and 72 hours for FB15K-237. We kept all hyperparameters the same as in the original paper, except for the reactive baseline constant $\lambda$, the regularization constant $\beta$, and the learning rate. The learning rate was $10^{-3}$ in all cases, and the remaining hyperparameters are shown in Table~\ref{hyperparameter}.

\paragraph{MultiHopKG baseline}
Tests were conducted on an Nvidia Tesla V100 GPU and took ~18 hours for NELL-995 and WN18RR and ~36 hours for FB15K-237. We did not alter any hyperparameters from their values in the original paper except for the learning rate, which we fixed at $10^{-3}$.

\subsubsection{Comparison results}
The evaluation results are shown in Table~\ref{performance comp} and Table~\ref{performance compfb60k}. The results of NeuralLP and M-Walk are quoted from \cite{minerva} and \cite{shen2018m}. The results of MINERVA and MultihopKG are regenerated using the code released by the corresponding authors\footnote{MINERVA: \url{https://github.com/shehzaadzd/MINERVA}; MultihopKG: \url{https://github.com/salesforce/MultiHopKG}. } on the best hyperparameter settings reported by them. 

From Table~\ref{performance comp} and Table~\ref{performance compfb60k}, we can see that our SSRL method consistently outperforms its baseline model for all metrics, i.e., SS-MINERVA outperforms MINERVA and SS-MultihopKG outperforms MultihopKG. Furthermore, our SSRL method (either SS-MINERVA or SS-MultihopKG) achieves the state-of-the-art on all datasets. Although NeuralLP gets the highest score on Hits@10 on WN18RR, we still claim our SSRL outperforms NeuralLP since SSRL gets higher scores on the other three metrics. Our SSRL does not perform the best on Hits@10 for WN18RR due to the performance gap between our baseline models and NeuralLP. 

\begin{figure*}[h]
\centering
    \includegraphics[scale=0.48]{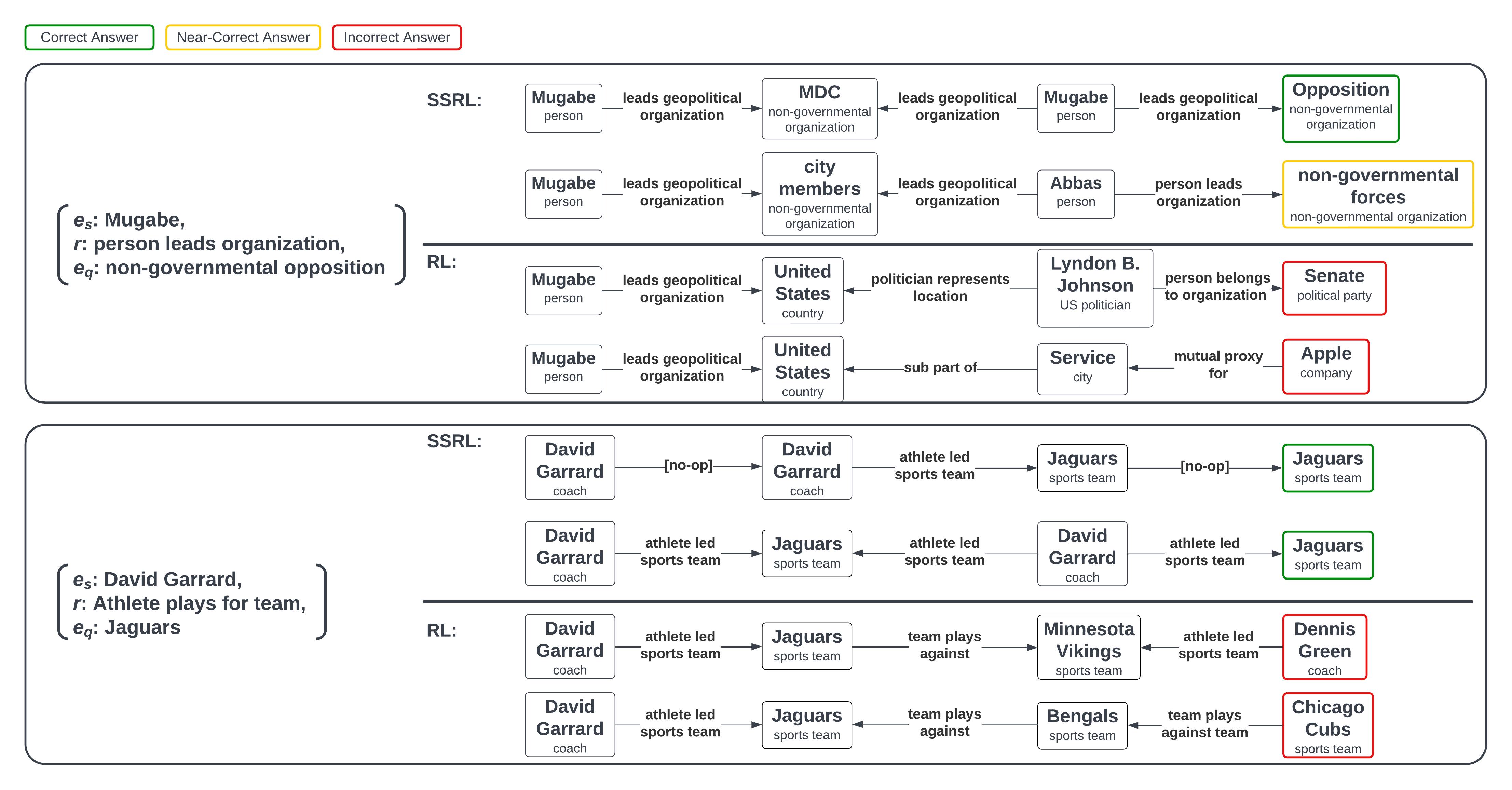}
    \caption{Reasoning paths discovered by the SSRL and RL-only agents respectively for 2 queries from the test set. A green box around the end entity indicates that the agent found the target entity exactly, a yellow box indicates a close match, and a red box indicates an incorrect entity.}
    \label{fig:path summary}
\end{figure*}

\subsubsection{Example paths}
Example reasoning paths that demonstrate the differences between SSRL and RL agents are presented in an interpretable form in Fig.~\ref{fig:path summary}. Overall, we find that the SSRL agent is much better at finding intuitive reasoning paths than the pure RL agent, arriving at correct or nearly correct answers consistently and taking appropriate paths to get there. 
For the example in Fig. \ref{fig:path summary} where $e_s =$ Mugabe, $r = $ person leads organization, and $e_q =$ nongovernmental opposition, the SSRL agent consistently picks the relation "leads geopolitical organization" analogous to $r$, arriving at correct or nearly correct (same category and conceptually similar) answers. In contrast, the RL agent appears only find organizations in general, and the relations it takes quickly diverge from $r$. 

The failures of the RL agent appear to be due to the lack of ability to use the context of $e_s$ in conjunction with the context of $r$. The RL agent knows to look for an organization, finding Apple and the Senate, but does not understand based on starting at Mugabe what kinds of organization to look for. In contrast, the SSRL agent travels through related concepts such as the MDC (a political party Mugabe led) and Abbas (President of Palestine and chairman of the NGO Fatah, which has fielded its own military forces). This is repeated in the second example where $e_s =$ David Garrard, $r = $ athlete plays for team, and $e_q =$ Jaguars, the SSRL agent either arrives and then stops early at the correct answer or circles back to it while the RL agent immediately overshoots the correct answer and fails to circle back. From this example, we can see that the RL agent understands to look for "athlete" and "sports team" based on $r$, but does not infer from starting with David Garrard (an NFL player who played for the Jaguars for most of his career) to narrow its search to football teams.

In short, both the RL and SSRL agents develop good understanding of what the relations mean in isolation, but only the SSRL agent develops a meaningful understanding of the relations in context of connecting two nodes in the knowledge base. We believe this is down to the information density of our SL pretraining method, which at each step gives the information about every action that can lead the agent to a correct answer, thus allowing the SSRL agent to get a better sense of how the relations it is learning work in a broader graph context.

\subsection{RL and SL comparison}
\begin{figure*}[h] 
    \centering
    \includegraphics[scale=0.12]{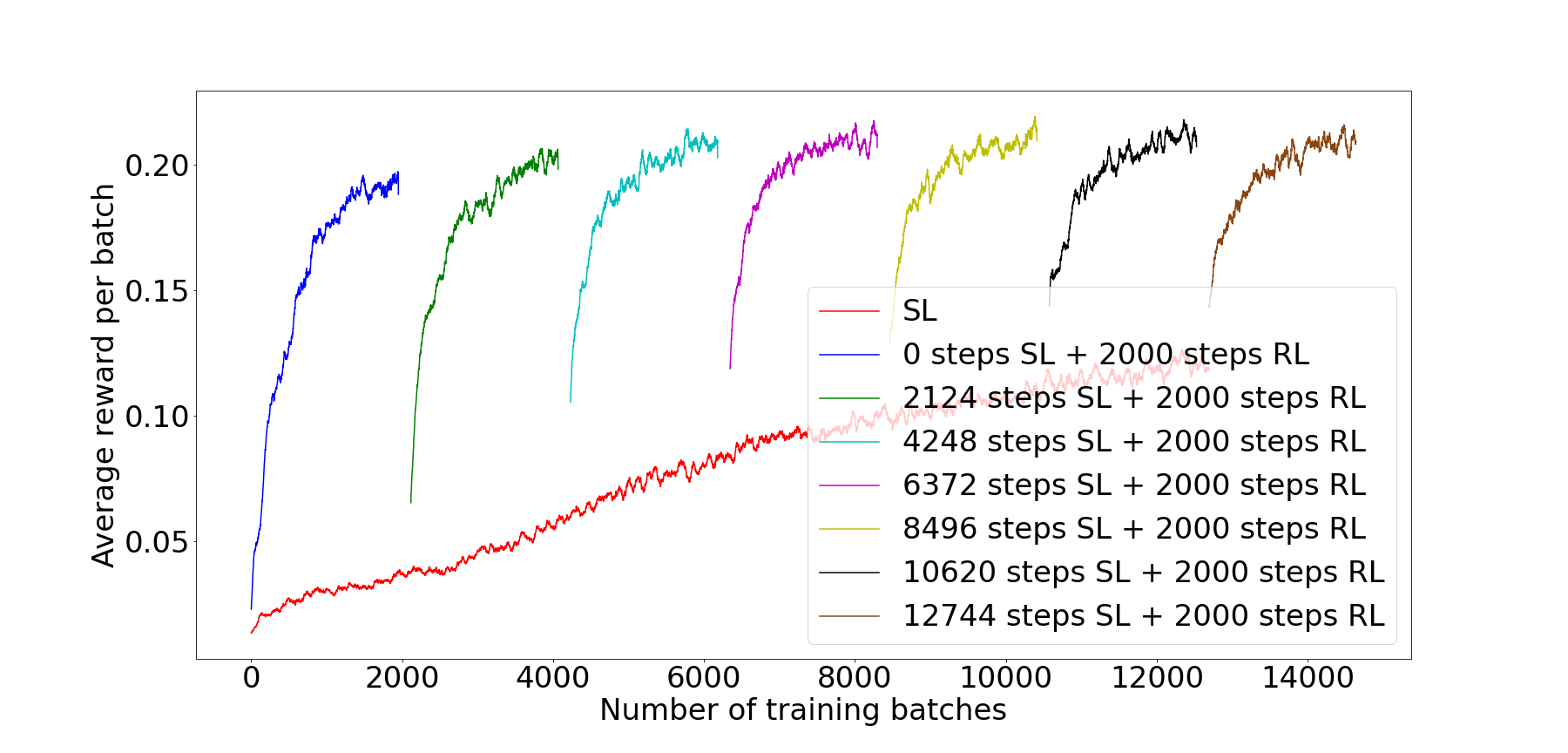}\hspace{-10mm} 
    \includegraphics[scale=0.12]{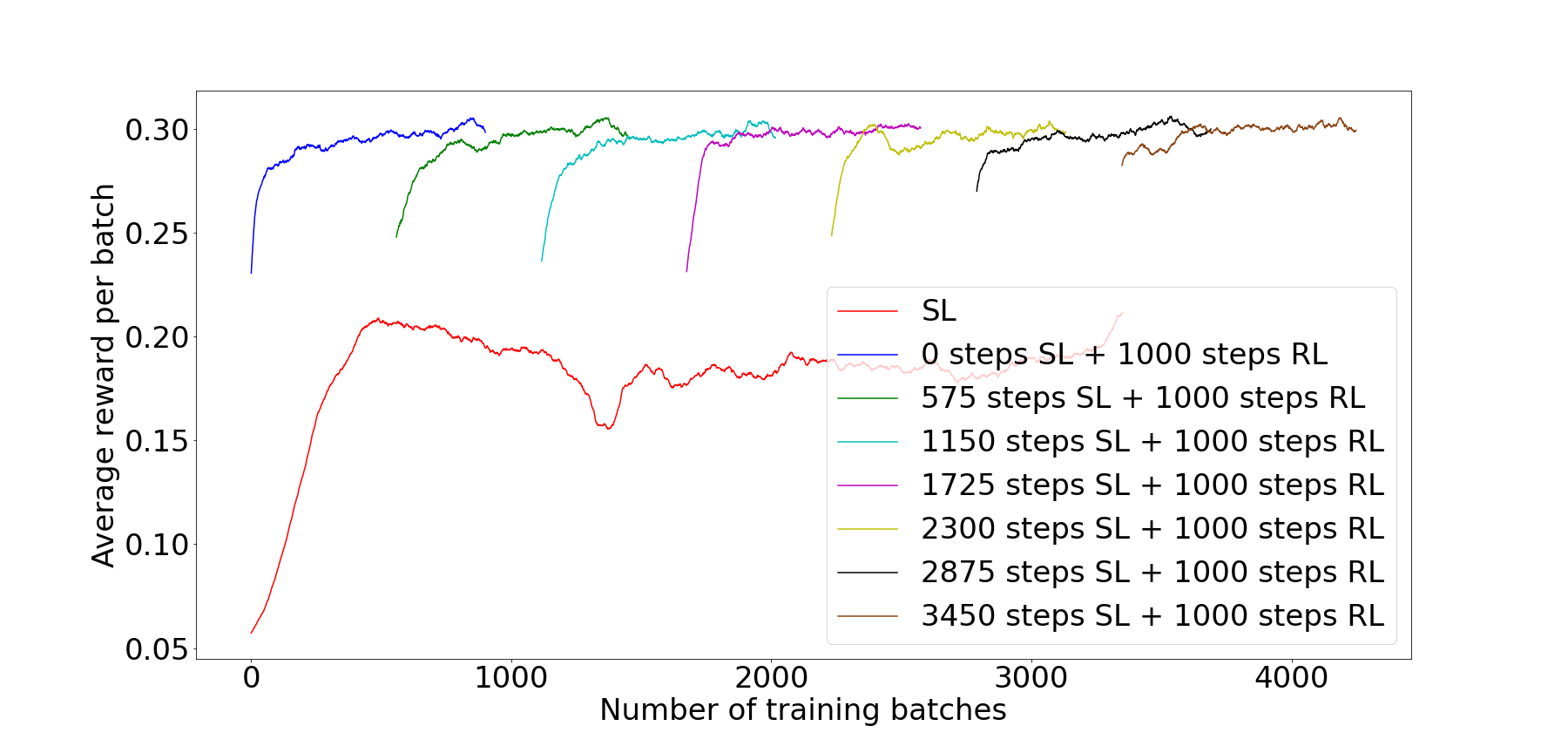}\\
    (a) FB15K-237  \hspace{45mm}  (b) WN18RR\\
    \includegraphics[scale=0.12]{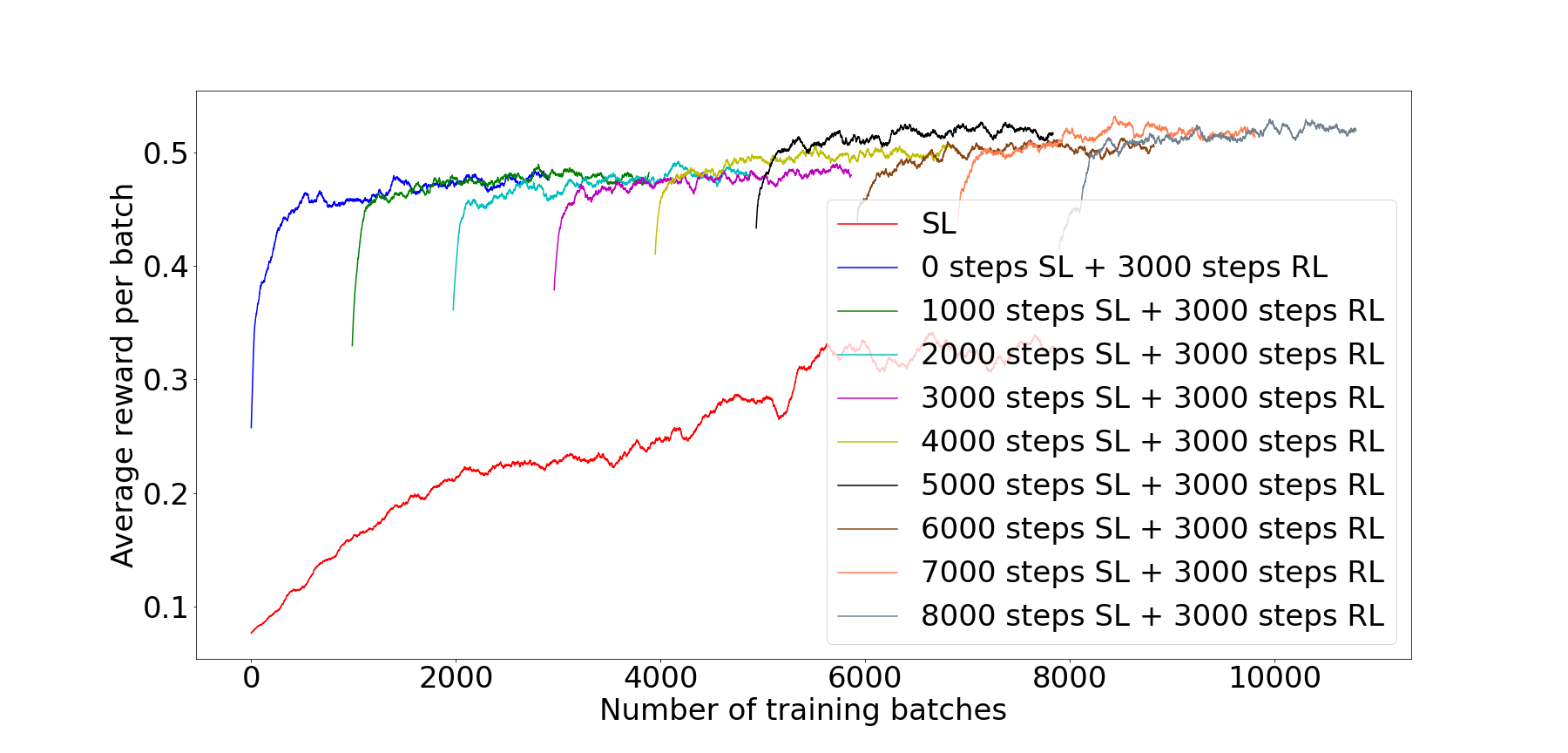}\hspace{-10mm} 
    \includegraphics[scale=0.12]{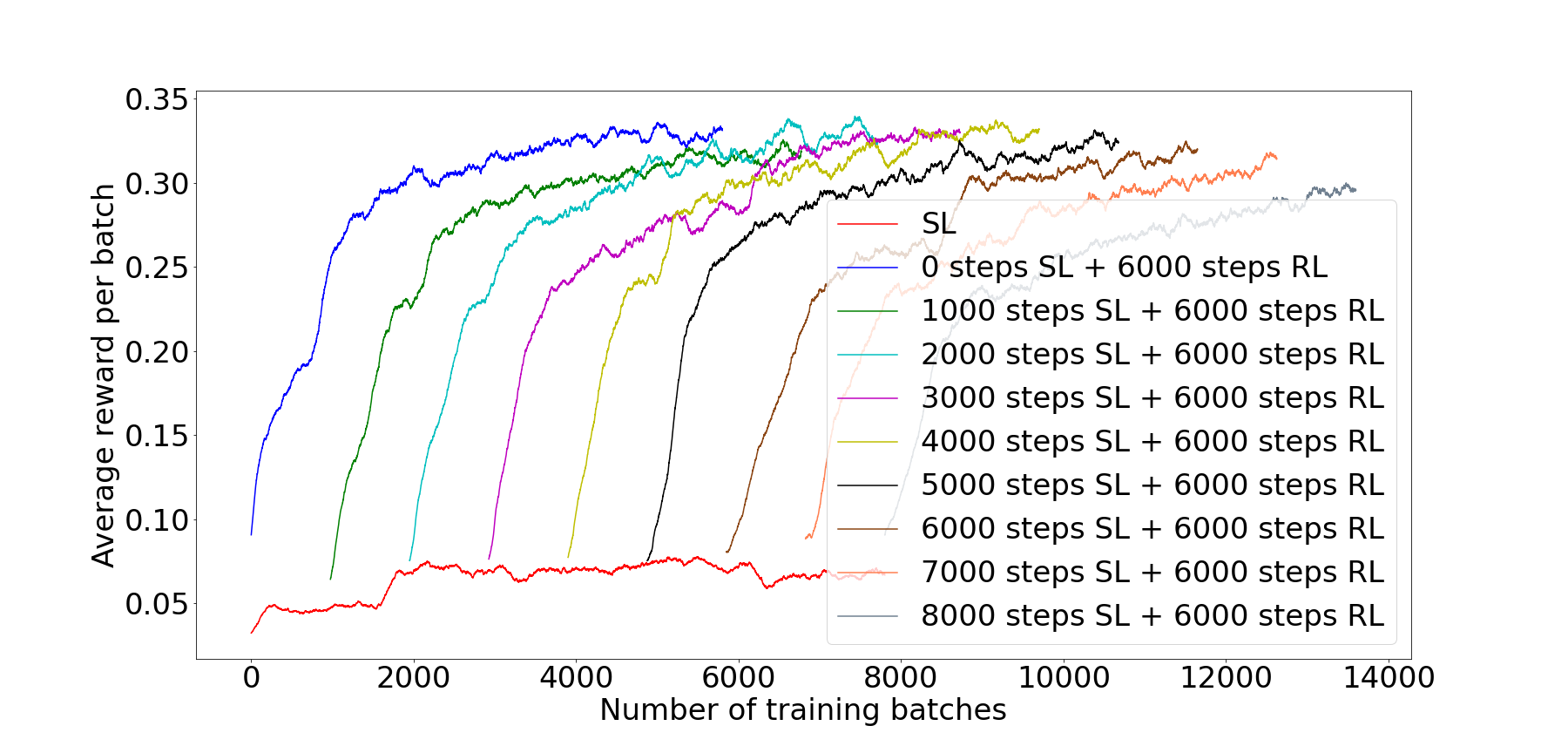}\\
    (c) NELL-995 \hspace{45mm} (d) FB60K
    \caption{Learning curves (accuracy v.s. number of training batches) with different SL pretraining steps followed by RL.}
    \label{fig:learing curve}
\end{figure*}

In this section, we verify our arguments about the pros and cons of SL and RL in terms of speed, coverage, and feasibility explained in Section \ref{train framework}.

\textbf{Speed}
We plot the learning curves of different SL training steps followed by a RL agent in Fig.~\ref{fig:learing curve}. From the figure, we can see that the RL agent learns much faster than SL. It is because the goal of SL is to learn the statistics of the whole underlying environment, which is much broader than the goal of RL agent, i.e., to find at least one path from the start entity to the end entity.
\begin{figure}
    \centering
    \includegraphics[scale=0.49]{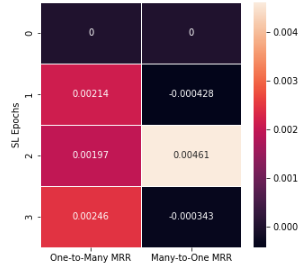}
    \hspace{-5mm}
    \includegraphics[scale=0.49]{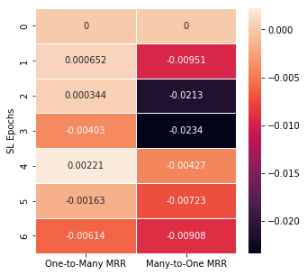}
    (a) FB15K-237\hspace{18mm} (b) WN18RR
                                           
    \caption{MRR evaluation on to-many and to-one queries.}
    \label{fig:onemany}
    
\end{figure}

\textbf{Coverage}
\begin{figure*}[h] 
    \centering
    \includegraphics[scale=0.4]{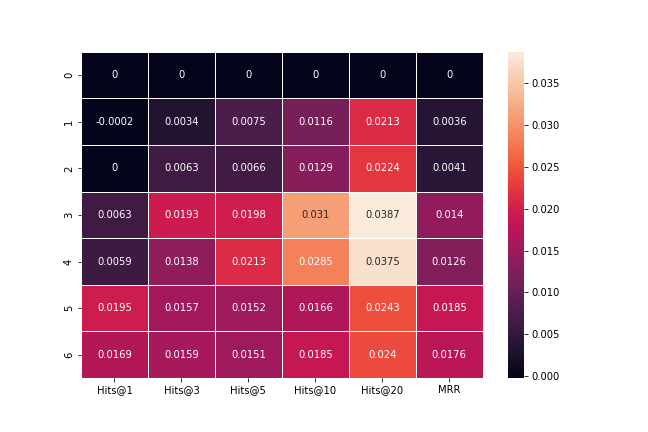}\hspace{-15mm}
    \includegraphics[scale=0.4]{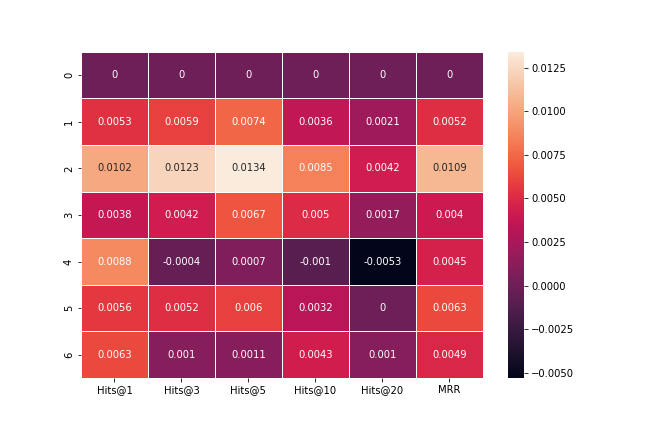}\hspace{-15mm}\\
    (a) FB15K-237  \hspace{50mm}  (b) WN18RR\\
        \includegraphics[scale=0.4]{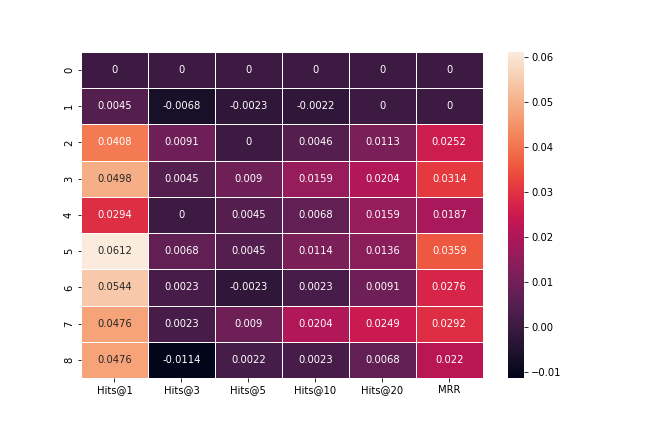}\hspace{-15mm}
    \includegraphics[scale=0.4]{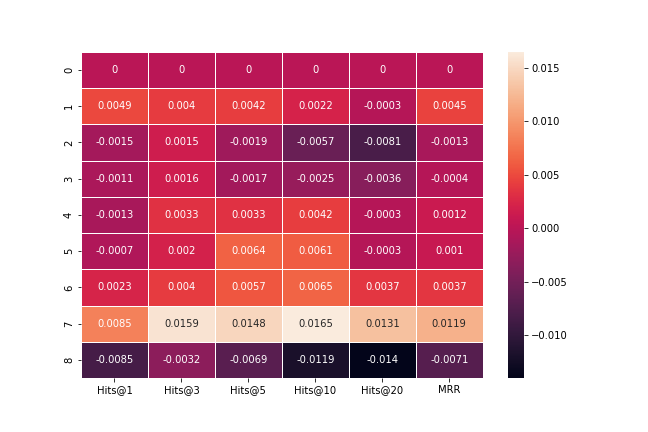}\\
    (c) NELL-995 \hspace{50mm} (d) FB60K
    \caption{Heatmap of Hits@k and MRR metrics for different SL training epochs followed by RL training (We show the version with MINERVA as baseline (SS-MINERVA)). Row 0 is pure RL. The value in each cell is the score for that cell minus the score for that category at epoch 0 to highlight the difference. The agent achieves its best performance with three, two, five, and seven epochs of SL pretraining on FB15K-237, WN18RR, NELL-995 and FB60K, respectively. The results reported in Tables \ref{performance comp} and \ref{performance compfb60k} are based on these SL training epochs. }
    \label{fig:heatmap}
\end{figure*}
Although SL-alone cannot achieve comparable results as RL in terms of both speed and performance, pretraining with SL helps the agent avoid time consuming exploration for queries with large action space. In other words, the SL pre-training stage helps the RL agent find the correct paths for difficult queries by expanding its coverage among all queries. 

In order to verify this argument, we compare the performance of our SSRL algorithm for to-many (there are more than one correct answers, $|E_{\mathbbm{all}}|>1$) and to-one queries (there is only one correct answer, $|E_{\mathbbm{all}}|=1$). Generally, the action space for to-many queries is larger than for to-one queries. These queries also have greater numbers of possible paths to a correct answer, which translates to more useful information on the SL labels and better performance. Fig.~ \ref{fig:onemany} shows the MRR values on to-many query set and to one query set. From the results, we can clearly see that SL are more helpful for to-many queries.

Furthermore, finding all paths appears crucial to finding a correct path, which aligns with the motivation behind our SL method of exposing the agent to a correct label of all valid actions, including those it is less likely to take, at each time step. For queries for which the agent found the correct answer during its beam search at the test time, the agent explored an average of 433 unique paths, while the agent explored only 61 unique paths in the opposite case. This highlights the intuitive conclusion that more well-connected $\left(e_s, e_q\right)$ pairs are easier to find paths for, but also highlights the fact that there is a strong correlation between the number of paths explored and the probability of finding an answer. If there was no benefit to finding more correct paths, then more or less well connected $\left(e_s, e_q\right)$ pairs wouldn’t be easier or harder to find paths for based on their connectedness.

We also need to point out that the objective functions for SL and RL are different. Increasing the SL training steps may not always improve performance. We show the Hits@k and MRR metrics for different SL pre-training epochs in Fig.~\ref{fig:heatmap}. From the results we can see that increasing the SL pre-training steps usually improves the performance on all metrics and then decreases it. As SL training progresses to a certain point, the differences in objective begin to hinder the agent when the switch to RL occurs. For example, the agent in RL achieves its best performance with three SL epochs before training on FB15K-237 and two epochs before training on WN18RR.

\textbf{Feasibility}
\begin{table}
\caption{Percentage of label used with different SL training epochs
}
\centering
\begin{tabular}{ccccccc}
\hline
Epoch & FB15K-237 &WN18RR & NELL-995 &FB60K\\
\hline
1 & 61.0\% & 81.6\% & 100.0\%  & 21.3\% \\
2 & 84.8\% & 96.6\% & 100.0\%  & 37.9\% \\
3 &  94.0\%  & 99.5\% & 100.0\% & 51.0\%  \\
4 &  97.7\%& 99.9\%& 100.0\%  & 61.4\% \\
5 &  99.1\% & 100.0\%& 100.0\%  & 69.6\% \\
6 &  99.6\% & 100.0\%& 100.0\%  & 69.6\% \\
7 &  -& -& 100.0\%  & 76.0\% \\
8 &  - & -& 100.0\%  & 81.0\% \\
\hline
\end{tabular}

\label{label usage}
\end{table}
Since generating labels for all training data is infeasible, we only use partial labels for the SL training stage. Table~\ref{label usage} shows the percentage of labeled data that we used among the entire training set. The results in Table~\ref{performance comp} and Table~\ref{performance compfb60k} are also based on this partial label pre-training. From the results we can draw the conclusion that pretraining with partial labels is conductible for these four large datasets.

\subsection{Performance difference on different datasets}
\label{sec:appendix}
\begin{figure*}[h!]
    \centering
    \includegraphics[scale=0.22]{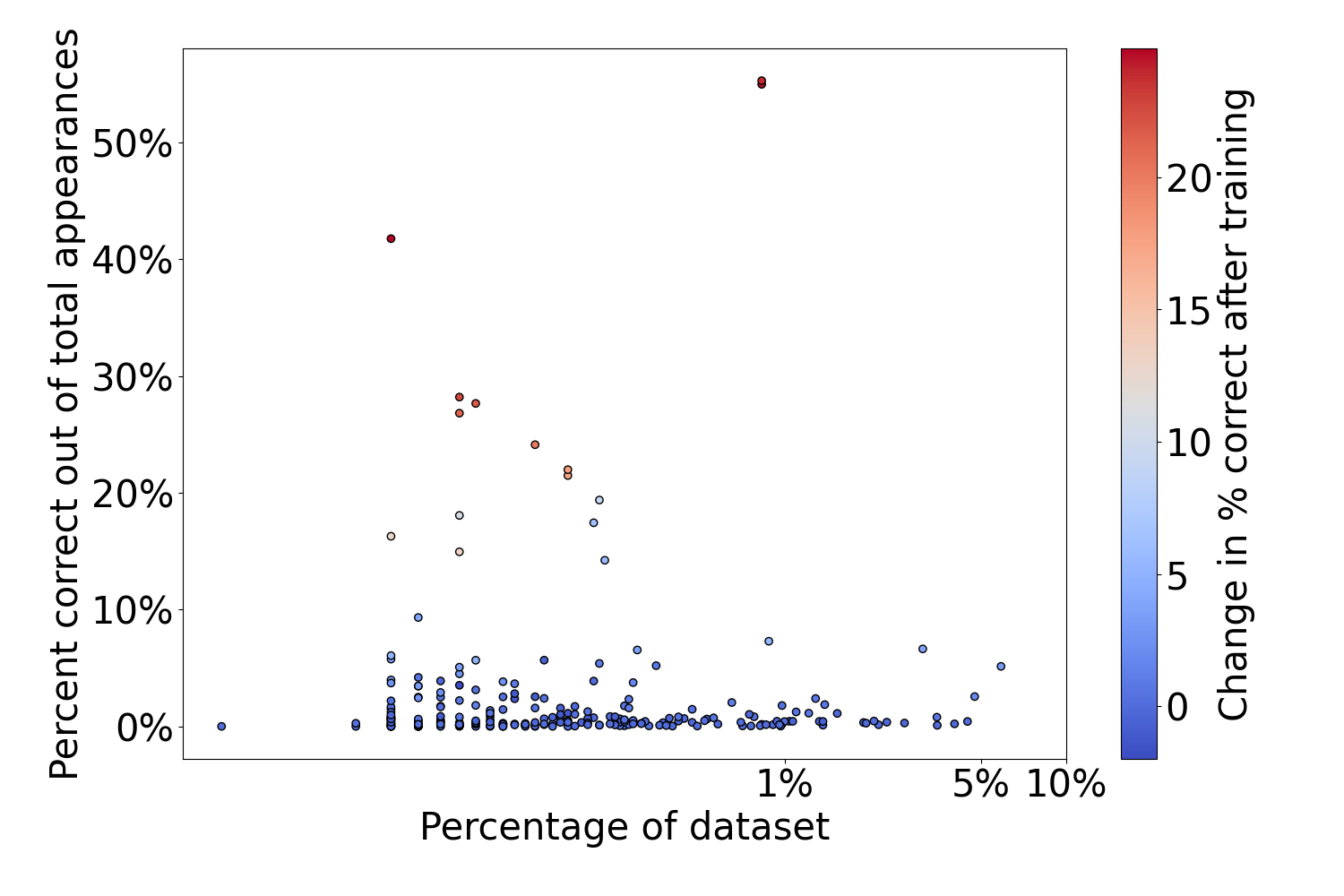}
    \includegraphics[scale=0.22]{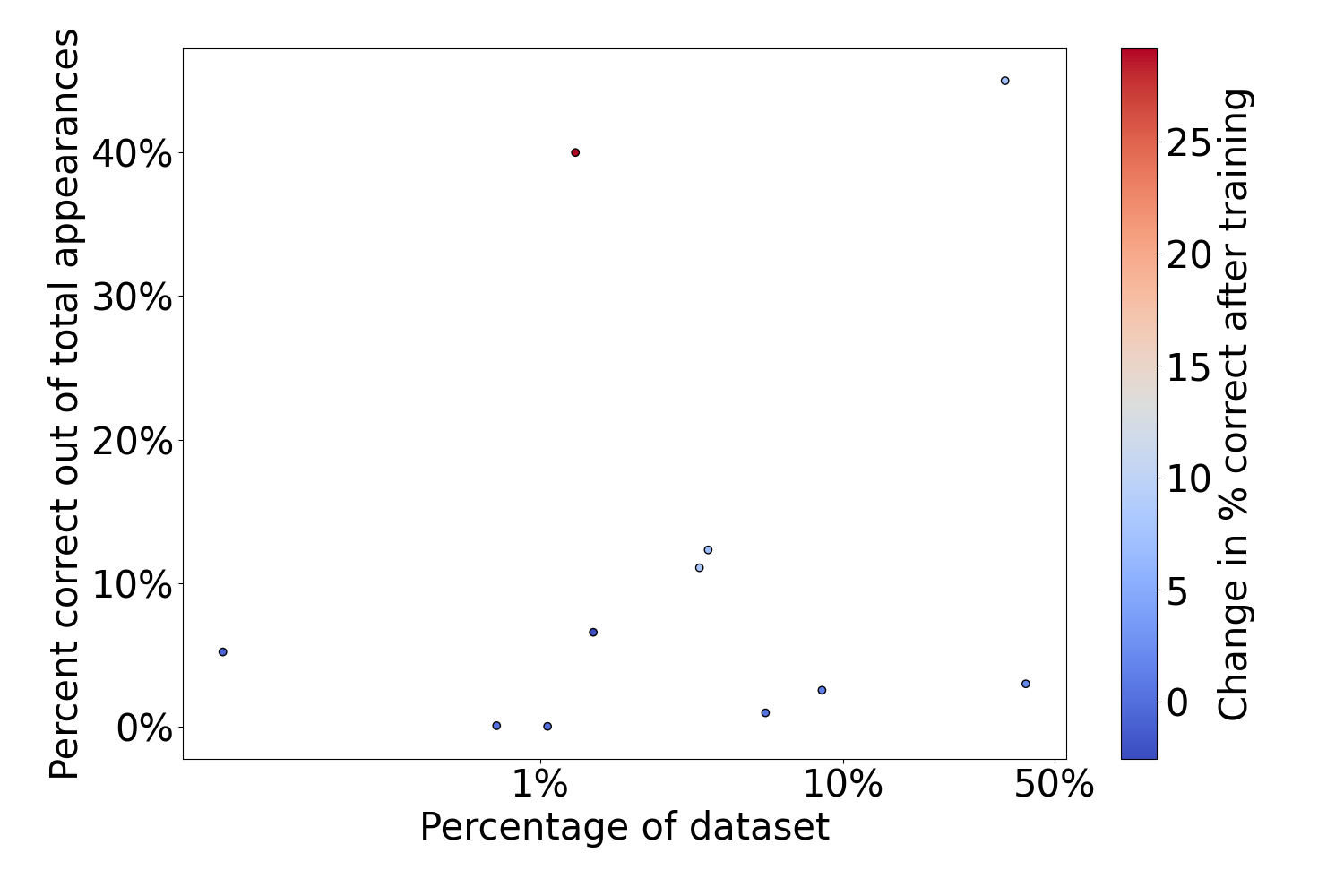}\\
    (a) FB15K-237  \hspace{50mm}  (b) WN18RR\\
    \includegraphics[scale=0.22]{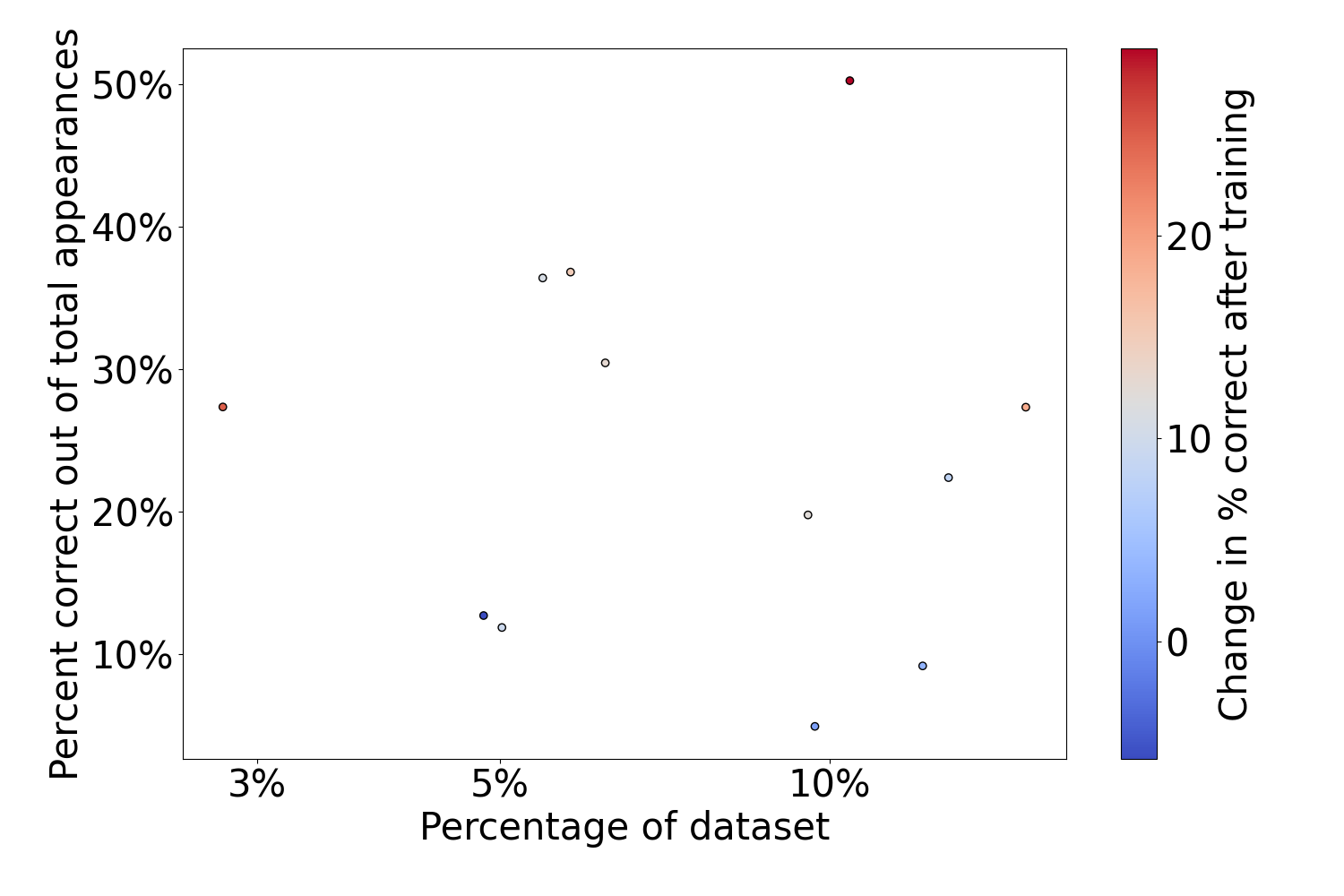}
    \includegraphics[scale=0.22]{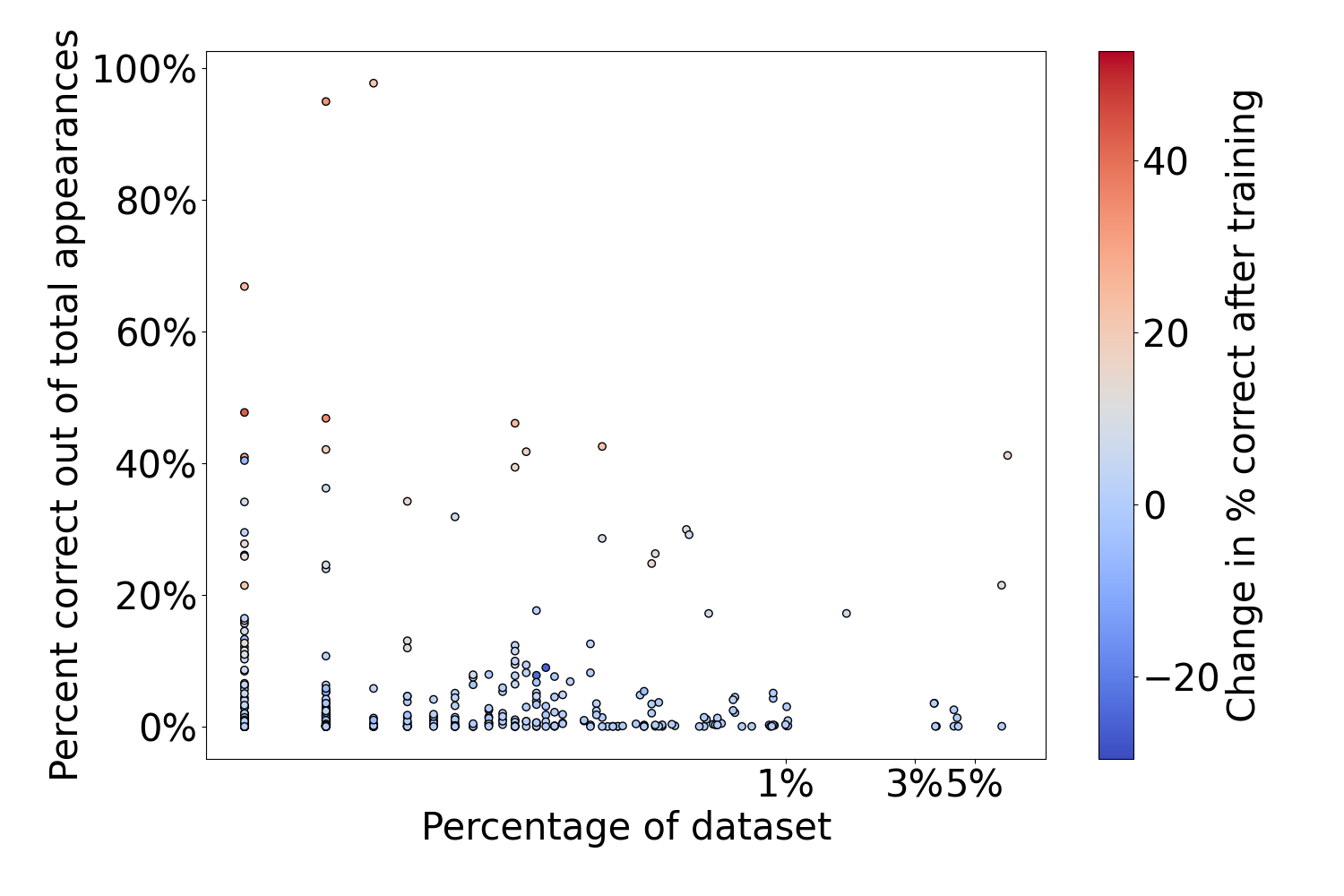}\\
    (c) NELL-995 \hspace{50mm} (d) FB60K
    \caption{Distribution of relations in the training set and final correct response percentage. Distribution is plotted on a logarithmic scale. The colors correspond to how much the correct response percentage for the relation increased over the course of training.}
    \label{fig:relation distribtion}
\end{figure*}

\begin{figure*}[h!]
    \centering
    \includegraphics[scale=0.13]{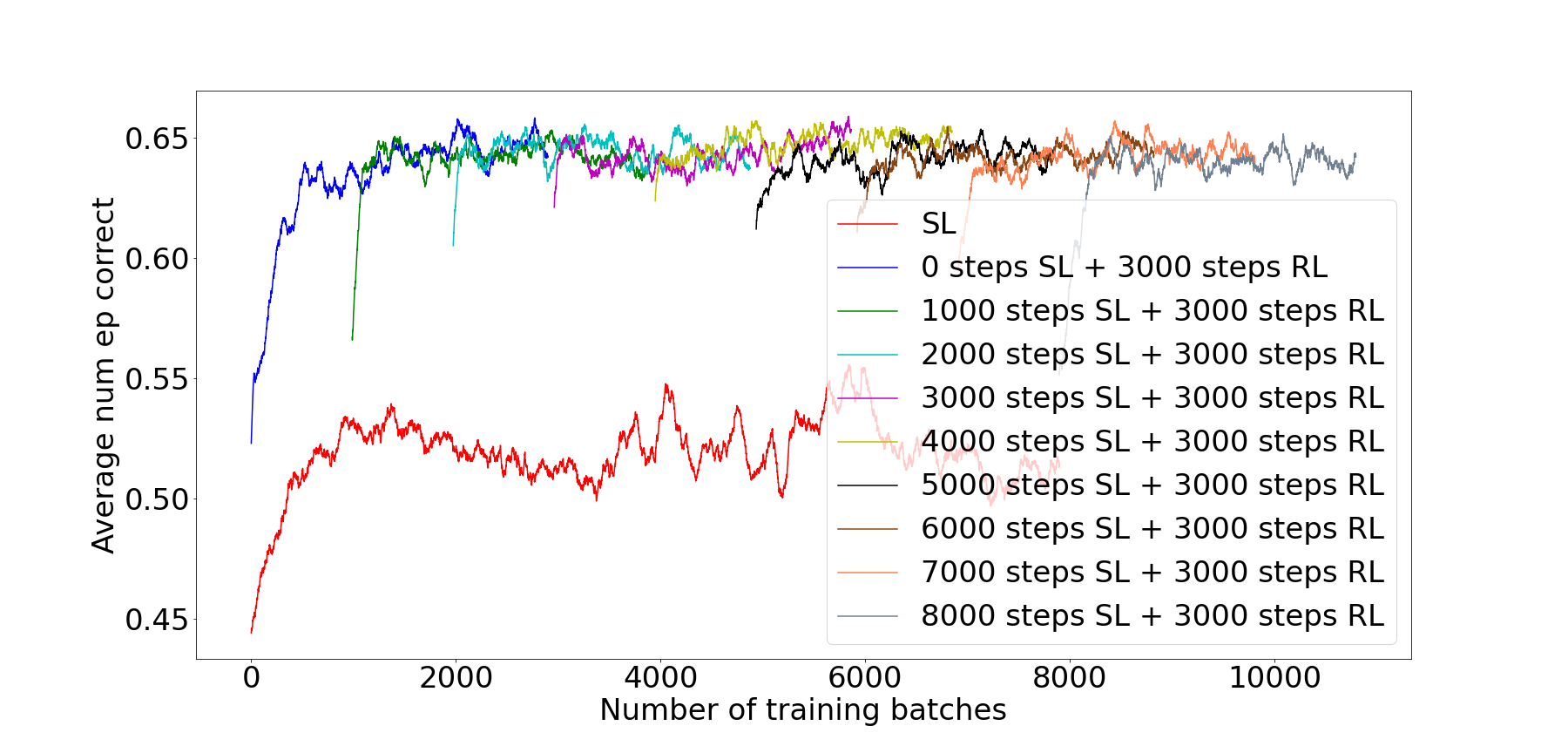}\hspace{-10mm} 
    \includegraphics[scale=0.13]{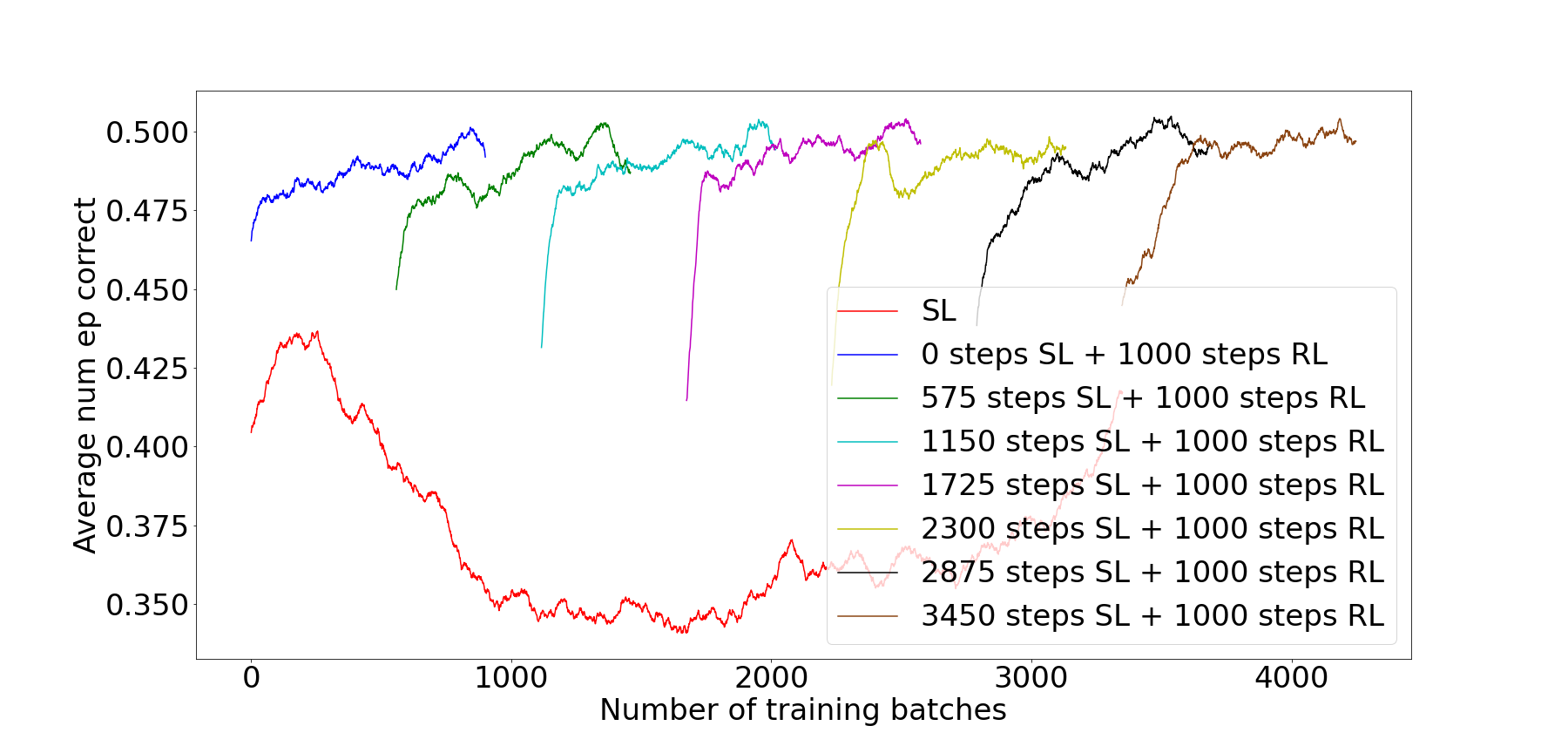}\\
    (a) FB15K-237  \hspace{45mm}  (b) WN18RR\\
    \includegraphics[scale=0.13]{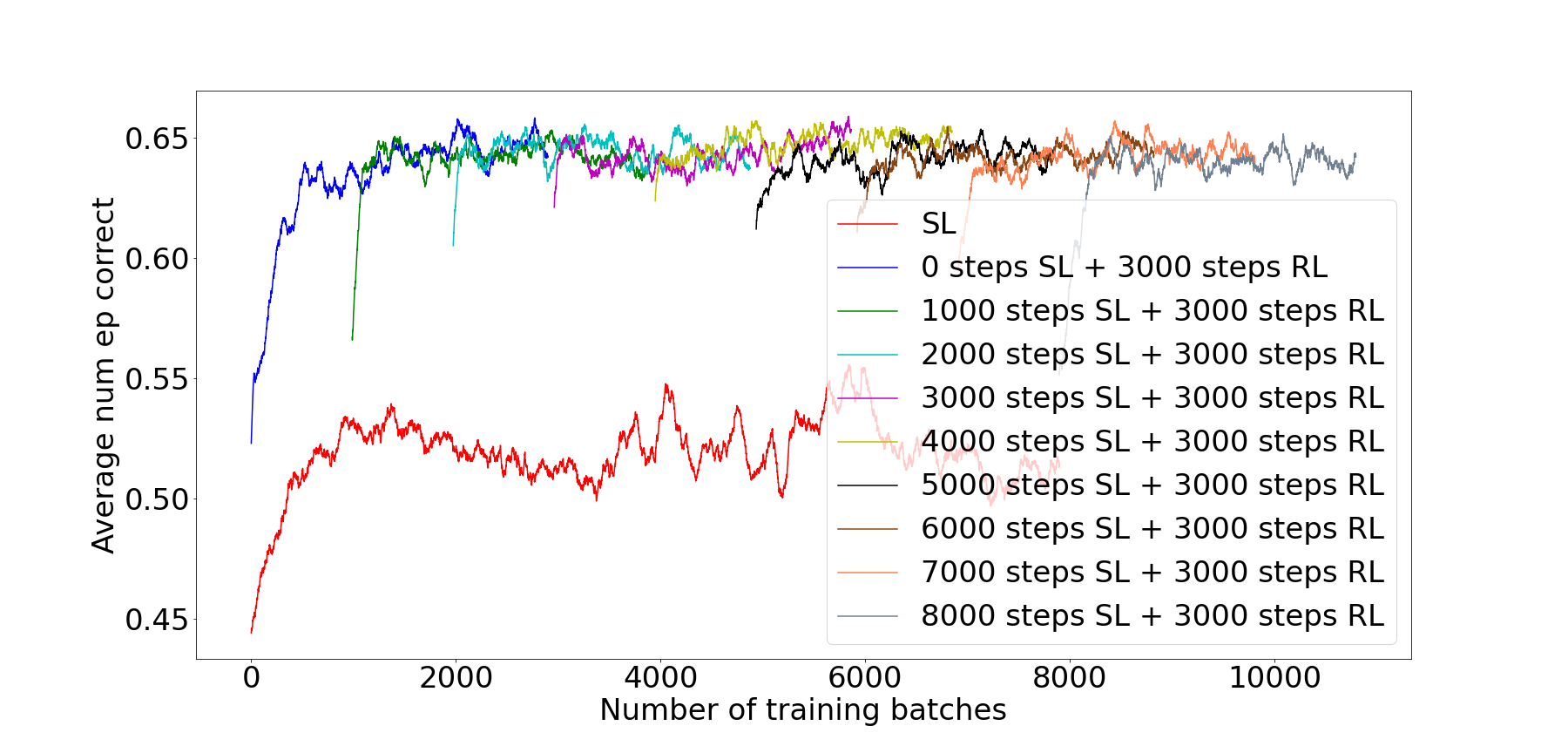}\hspace{-10mm} 
    \includegraphics[scale=0.13]{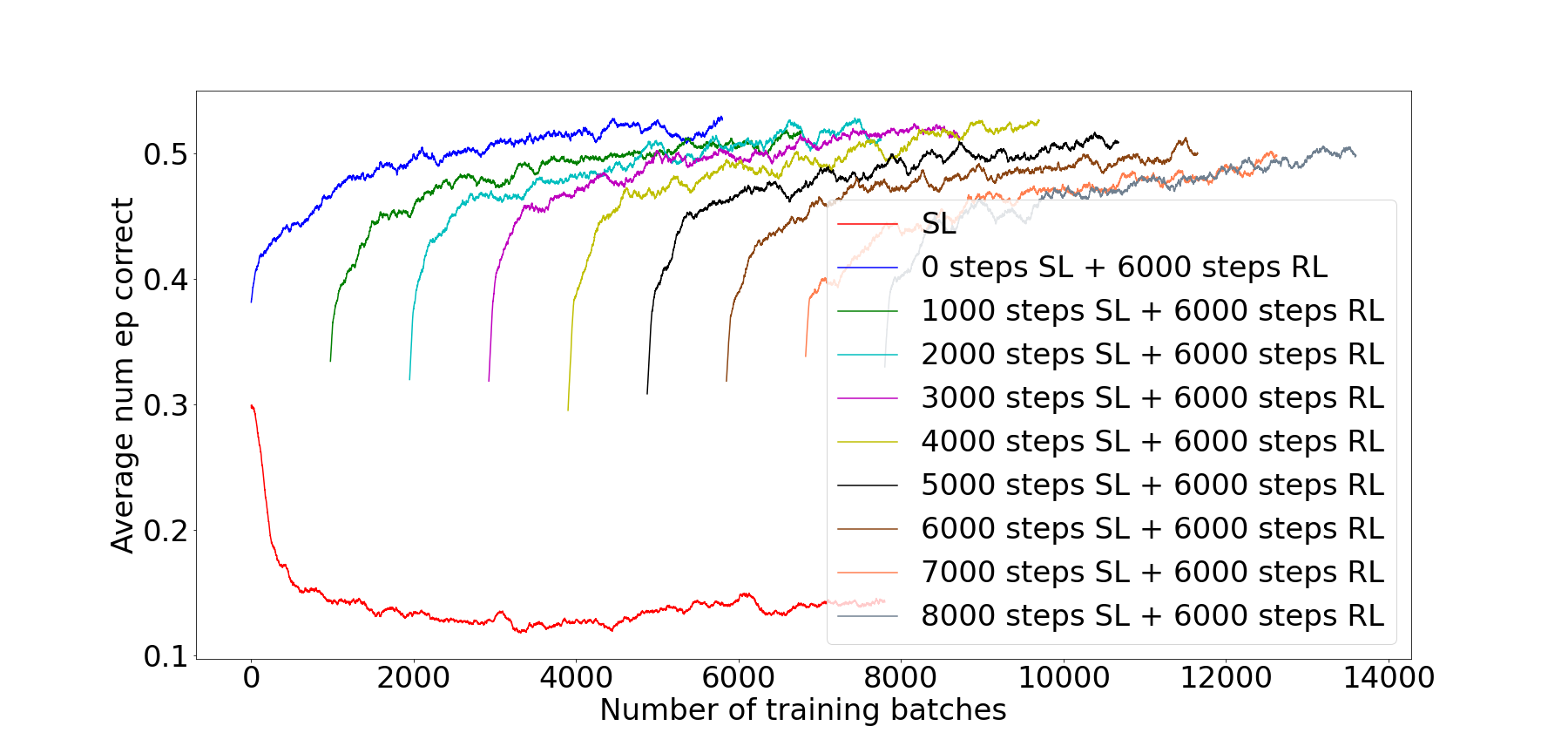}\\
    (c) NELL-995 \hspace{45mm} (d) FB60K
    \caption{Learning curves (Hits@20 v.s. number of training batches) with different SL pretraining steps followed by same RL steps}
    \label{fig:learing curve2}
\end{figure*}

From Table \ref{performance comp} and Table \ref{performance compfb60k}, we can see that the improvements of SSRL to RL on different KG datsets are quite different. The improvement on NELL-995 datasets are significant. However, the improvement in WN18RR and FB60K is only slight. We think there are two reasons for it: 
\begin{itemize}
    \item The imbalanced training data for different relation types.
    \item The small graph degree (number of edges connecting to each edge);
\end{itemize}

In order to verify these two assumptions, we show the distribution of relations in the training set in Fig.~\ref{fig:relation distribtion}. On NELL, where the agent performs the best on out of all the datasets, there is an even distribution of relation types that show up with different frequencies in the dataset. This means that the agent is seeing a good sample of the different query types and trains well on all of them. On WN18RR and FB60K, the distributions of relation types are very skewed. 

For WN18RR, where the improvement of SSRL on RL is the smallest, only two types of relations comprise more than 75\% of the dataset while the others appear comparatively infrequently. Since the goal of SL is to learn the underlying distribution based on labeled data and encourage the agent to explore paths it would not otherwise, pretraining on these skewed distributions would encourage the agent to find more correct paths on the dominate relations, so it achieves higher overall accuracy on the whole training set. However, it may sacrifice performance on less common relations, where finding even one correct path is difficult. Furthermore, the idea behind SL is that by teaching the agent this underlying distribution it may take potentially advantageous paths it otherwise would not take; if the the dataset is mostly homogeneous (e.g. sharply skewed towards a select few relations), there is little that SL could show the agent that it would not find during RL stage. As a result, SL pretraining does not help in terms of Hits@k when k is large on sharply skewed datasets like WN18RR. 

For FB60K, Fig.~\ref{fig:heatmap} shows that the trial in which a correct path was found for the highest percentage of queries uses 7000 SL steps. The graphs in Fig.~\ref{fig:learing curve} in the main manuscript track the average reward throughout the batch, which includes multiple trials for each query. Our analysis of agent performance by relation type included in Fig.~\ref{fig:relation distribtion} shows that FB60K is the only dataset containing numerous relations with a near 100\% fact prediction success rate, implying that it contains facts that are easier for the agent to understand. Intuitively, the immediately strong performance on these "low hanging fruit" will increase the average performance towards the beginning of training, but as increasing durations of SL pretraining force the agent to generalize to more complex relations, its success rate on these relations naturally falls, resulting in a decreasing average performance and an increasing percentage of queries for which a correct path is found. This matches the graph and heatmap presented in Fig.~\ref{fig:learing curve} and Fig.~\ref{fig:heatmap}.

To back up these claims, we plot the learning curves with Hit@20 as the metric in Fig.~\ref{fig:learing curve2}. SL graphs show negative trends on skewed datasets, i.e., WN18RR and FB60K. This is reasonable because the target of SL is to minimize the crossentropy loss and therefore maximize the overall accuracy. It is noted that the final performance on Hits@20 can be fixed by the RL training stage. This also shows the importance of the RL training stage in the proposed architecture.

We also observed that our SSRL architecture performs well on FB15K-237, where the relation-type distribution is also skewed (although it is less skewed than FB60K and WN18RR). We believe the excellent performance is due to the high graph degree, where the SL are particularly helpful. Table 1 of the main manuscript shows that the median degree of FB15K-237 is 14 whereas other datasets are less than 4.

In summary, the extent of improvement of SSRL to RL depends on the statistics of the datasets such as graph degree and relation distributions. Currently, our SSRL architecture generates labels for random relation types. In our future work, we will discuss how to generate labels based on the relation distributions in each dataset, therefore improving the performance further.

\subsection{Comparison with DeepPath}
\label{deeppath comp}
\begin{table}
\centering
\caption{Comparison of mean average precision and SR (success rate) for DeepPath with its original SL pretraining method and DeepPath with our SL pretraining method for the link prediction.
}
\begin{tabular}{lccc}
\hline
Data &Metric &Ours&DeepPath\\
\hline
\multirow{2}{5em}{FB15K-237} & MAP &\textbf{48.5}&46.9\\
& SR &\textbf{0.055}&0.029\\

\hline
\multirow{2}{5em}{NELL-995} & MAP &\textbf{0.73} &0.663\\
& SR&\textbf{0.127} &0.057 \\

\hline

\end{tabular}

\label{performance comp with deeppath}
\end{table}

\begin{table}
\centering
\caption{Comparison of mean average precision scores of DeepPath with its original SL pretraining method and DeepPath with our SL FB15K-237 dataset
}
\begin{tabular}{lp{0.018\textwidth}cp{0.018\textwidth}c}
\hline
\multirow{2}{4em}{Relation} & \multicolumn{2}{c}{MAP}  &\multicolumn{2}{c}{SR} \\
&Ours&DeepPath&Ours&DeepPath\\
\hline
PhoneServiceLocation& 0.31&0.49& 0.014& 0.01\\
CelebrityRomanticRelationship& 0.51&0.57 & 0.06&0.08\\
EducationalInstitutionCampus& 1&1&0.06&0.08\\
FilmDirector& 0.44 &0.43&0.05&0.04\\
FilmCinematography & 0.19&0.26&0.00&0.00\\
FilmFromCountry& 0.52&0.48 &0.03&0.02\\
FilmLanguage& 0.50&0.59 &0.004&0.006\\
FilmMusic& 0.29 &0.27&0.028&0.014\\
FilmWrittenBy & 0.56&0.56&0.14&0.08\\
PersonNationality&0.62&0.18&0.016&0.004\\
AdministrativeDivisionOfCapital&0.85 &0.83&0.47&0.25\\
LocationContains& 0.52 &0.51&0.02 &0.01\\
SymptomOfDisease& 0.48 &0.48&0 &0.007\\
ArtistOrigin&0.46&0.46&0.024&0.028\\
PersonFoundedOrganization&0.28&0.24&0.027&0\\
OrganizationHeadquartersCity&0.51&0.45&0.13&0.02\\
MemberOfOrganization&0.22&0.22&0.09&0.003\\
LeadsOrganization&1&1&0.003&0\\
CauseOfDeath&0.83&0.84&0.002&0\\
LanguagesSpoken&0.35&0.35&0&0.006\\
PersonNationality&0.62&0.18&0.016&0.004\\
PlaceOfBirth&0.47&0.51&0.178&0.032\\
IsReligion&0.23&0.23&0&0\\
MarriedTo&0.38&0.45&0.164&0.043\\
SpecializationOf&0.44&0.36&0.012&0\\
CityHasSportsTeam&0.47&0.43&0.003&0\\
playsFootballPosition&0.32&0.31&0.014&0.016\\
TeamPlaysSport&0.42&0.42&0&0\\
eventInLocation&0.37&0.35&0.007&0\\
TVProgramFromCountry&0.79&0.65&0.022&0.027\\
TVProgramGenre&0.18&0.18&0.004&0\\
TVProgramLanguage&0.53&0.45&0.047&0.009\\
\hline
\textbf{Mean}&\textbf{0.49}&0.47&\textbf{0.055}&0.029\\
\hline
\end{tabular}

\label{performance compdeeppath fb15k}
\end{table}

We also evaluate the effectiveness of our SL pre-training method on the link prediction task solved by DeepPath, in comparison to their own SL pre-training. Since DeepPath tests each relation separately, to make a fair comparison, we modify our SL pretraining algorithm to use a normalized histogram of number of correct paths per relation (as the space of relations is the action space) at the current state and substitute it for DeepPath's original SL pretraining method during that portion of training.

The results of these tests on the NELL-995 and FB15K-237 datasets are presented in Table~\ref{performance comp with deeppath} and the performance for each relation is shown in Table ~\ref{performance compdeeppath fb15k} and Table ~\ref{performance compdeeppath nell}. We believe the superior performance derived from our method is due to two key differences in our methods: the information density of our SL objective and the freedom of exploration we grant the agent.

\begin{table}
\centering

\caption{Comparison of mean average precision scores of DeepPath with its original SL pretraining method and DeepPath with our SL pretraining method for the link prediction task on the NELL-995 dataset}
\begin{tabular}{lp{0.018\textwidth}cp{0.018\textwidth}c}
\hline 
\multirow{2}{4em}{Relation} & \multicolumn{2}{c}{MAP}  &\multicolumn{2}{c}{SR} \\
&Ours&DeepPath&Ours&DeepPath\\
\hline
AgentBelongsToOrganization&0.577&0.580&0.220&0.194\\
AthleteHomeStadium&0.830&0.830&0.043&0.002\\
AthletePlaysForTeam&0.733&0.730&0.096&0.036\\
AthletePlaysInLeague&0.578&0.588&0.012&0.006\\
AthletePlaysSport&0.840&0.765&0.002&0\\
OrganizationHeadquarteredInCity&0.790&0.790&0.310&0.172\\
OrganizationHiredPerson&0.748&0.700&0.186&0.038\\
PersonBornInLocation&0.699&0.699&0.176&0.194\\
PersonLeadsOrganization&0.792&0.789&0.232&0.002\\
TeamPlaysInLeague&0.835&0.292&0.012&0.036\\
TeamPlaysSport&0.649&0.491&0.012&0.006\\
WorksFor&0.691&0.699&0.224&0\\
\hline
\textbf{Mean}&\textbf{0.730}&0.663&\textbf{0.127}&0.057\\
\hline

\end{tabular}

\label{performance compdeeppath nell}
\end{table}

The primary advantage of our method is that at each time step, the label compared with the agent's computed action probabilities contains contextualized information about every path which can be taken from the current state to the correct $e_q$ for the query. This information density allows the agent to learn more at each time step than if it was following a single randomly selected example path. Furthermore, these labels contain information about paths through entities with a low degree that would be less likely to appear in a path randomly sampled from the pool of correct paths. Overall, this allows the agent to learn about entities in the context of the graph as a whole rather than in isolation or in the context of a single query, allowing greater generalization.

The secondary advantage of our method is that the actions taken are selected by the agent instead of according to a pre-selected path. Selecting the actions for the agent inherently means that the agent is being trained in situations it might be unlikely to encounter while traversing the graph on its own, making the knowledge gained less applicable. In essence, our method is allowing the agent to work and correcting its misunderstandings as they reveal themselves, rather than simply showing it the correct answer and moving on.

\section{Conclusion}
In this paper, we propose a self-supervised RL agent that can warm up its parameters by using automatically generated partial labels. We discuss the pros and cons of pure SL and RL for KGR task and show experimentally that supervised pretraining followed by reinforcement training combines the advantage of SL and RL, i.e., the SSRL architecture achieves start of art performance on four large KG datasets and the SSRL agent consistently outperforms its RL baseline with all Hits@k and MRR metrics. To compare with the general SSRL framework, we adapt our SSRL pretraining method to link prediction task and proves the superiority of our SL strategy experimentally.

\bibliographystyle{IEEEtran}
\bibliography{ref}

\begin{thebibliography}{10}
\providecommand{\url}[1]{#1}
\csname url@samestyle\endcsname
\providecommand{\newblock}{\relax}
\providecommand{\bibinfo}[2]{#2}
\providecommand{\BIBentrySTDinterwordspacing}{\spaceskip=0pt\relax}
\providecommand{\BIBentryALTinterwordstretchfactor}{4}
\providecommand{\BIBentryALTinterwordspacing}{\spaceskip=\fontdimen2\font plus
\BIBentryALTinterwordstretchfactor\fontdimen3\font minus \fontdimen4\font\relax}
\providecommand{\BIBforeignlanguage}[2]{{%
\expandafter\ifx\csname l@#1\endcsname\relax
\typeout{** WARNING: IEEEtran.bst: No hyphenation pattern has been}%
\typeout{** loaded for the language `#1'. Using the pattern for}%
\typeout{** the default language instead.}%
\else
\language=\csname l@#1\endcsname
\fi
#2}}
\providecommand{\BIBdecl}{\relax}
\BIBdecl

\bibitem{yu-etal-2022-kg}
\BIBentryALTinterwordspacing
D.~Yu, C.~Zhu, Y.~Fang, W.~Yu, S.~Wang, Y.~Xu, X.~Ren, Y.~Yang, and M.~Zeng, ``{KG}-{F}i{D}: Infusing knowledge graph in fusion-in-decoder for open-domain question answering,'' in \emph{Proceedings of the 60th Annual Meeting of the Association for Computational Linguistics (Volume 1: Long Papers)}.\hskip 1em plus 0.5em minus 0.4em\relax Dublin, Ireland: Association for Computational Linguistics, May 2022, pp. 4961--4974. [Online]. Available: \url{https://aclanthology.org/2022.acl-long.340}
\BIBentrySTDinterwordspacing

\bibitem{he-etal-2017-learning}
\BIBentryALTinterwordspacing
H.~He, A.~Balakrishnan, M.~Eric, and P.~Liang, ``Learning symmetric collaborative dialogue agents with dynamic knowledge graph embeddings,'' in \emph{Proceedings of the 55th Annual Meeting of the Association for Computational Linguistics (Volume 1: Long Papers)}.\hskip 1em plus 0.5em minus 0.4em\relax Vancouver, Canada: Association for Computational Linguistics, Jul. 2017, pp. 1766--1776. [Online]. Available: \url{https://aclanthology.org/P17-1162}
\BIBentrySTDinterwordspacing

\bibitem{moon-etal-2019-opendialkg}
\BIBentryALTinterwordspacing
S.~Moon, P.~Shah, A.~Kumar, and R.~Subba, ``{O}pen{D}ial{KG}: Explainable conversational reasoning with attention-based walks over knowledge graphs,'' in \emph{Proceedings of the 57th Annual Meeting of the Association for Computational Linguistics}.\hskip 1em plus 0.5em minus 0.4em\relax Florence, Italy: Association for Computational Linguistics, Jul. 2019, pp. 845--854. [Online]. Available: \url{https://aclanthology.org/P19-1081}
\BIBentrySTDinterwordspacing

\bibitem{huang2019knowledge}
X.~Huang, J.~Zhang, D.~Li, and P.~Li, ``Knowledge graph embedding based question answering,'' in \emph{Proceedings of the twelfth ACM international conference on web search and data mining}, 2019, pp. 105--113.

\bibitem{minerva}
R.~Das, S.~Dhuliawala, M.~Zaheer, L.~Vilnis, I.~Durugkar, A.~Krishnamurthy, A.~Smola, and A.~McCallum, ``Go for a walk and arrive at the answer: Reasoning over paths in knowledge bases using reinforcement learning,'' in \emph{ICLR}, 2018.

\bibitem{shen2018m}
Y.~Shen, J.~Chen, P.-S. Huang, Y.~Guo, and J.~Gao, ``M-walk: Learning to walk over graphs using monte carlo tree search,'' \emph{Advances in Neural Information Processing Systems}, vol.~31, 2018.

\bibitem{LinRX2018:MultiHopKG}
X.~V. Lin, R.~Socher, and C.~Xiong, ``Multi-hop knowledge graph reasoning with reward shaping,'' in \emph{Proceedings of the 2018 Conference on Empirical Methods in Natural Language Processing, {EMNLP} 2018, Brussels, Belgium, October 31-November 4, 2018}, 2018.

\bibitem{wenhan_emnlp2017}
W.~Xiong, T.~Hoang, and W.~Y. Wang, ``Deeppath: A reinforcement learning method for knowledge graph reasoning,'' in \emph{Proceedings of the 2017 Conference on Empirical Methods in Natural Language Processing (EMNLP 2017)}.\hskip 1em plus 0.5em minus 0.4em\relax Copenhagen, Denmark: ACL, September 2017.

\bibitem{ravichandar2020recent}
H.~Ravichandar, A.~S. Polydoros, S.~Chernova, and A.~Billard, ``Recent advances in robot learning from demonstration,'' \emph{Annual review of control, robotics, and autonomous systems}, vol.~3, pp. 297--330, 2020.

\bibitem{toutanova-etal-2015-representing}
\BIBentryALTinterwordspacing
K.~Toutanova, D.~Chen, P.~Pantel, H.~Poon, P.~Choudhury, and M.~Gamon, ``Representing text for joint embedding of text and knowledge bases,'' in \emph{Proceedings of the 2015 Conference on Empirical Methods in Natural Language Processing}.\hskip 1em plus 0.5em minus 0.4em\relax Lisbon, Portugal: Association for Computational Linguistics, Sep. 2015, pp. 1499--1509. [Online]. Available: \url{https://aclanthology.org/D15-1174}
\BIBentrySTDinterwordspacing

\bibitem{cao2022geometry}
Z.~Cao, Q.~Xu, Z.~Yang, X.~Cao, and Q.~Huang, ``Geometry interaction knowledge graph embeddings,'' in \emph{AAAI Conference on Artificial Intelligence}, 2022.

\bibitem{chao2020pairre}
L.~Chao, J.~He, T.~Wang, and W.~Chu, ``Pairre: Knowledge graph embeddings via paired relation vectors,'' \emph{arXiv preprint arXiv:2011.03798}, 2020.

\bibitem{li2022does}
R.~Li, Y.~Cao, Q.~Zhu, G.~Bi, F.~Fang, Y.~Liu, and Q.~Li, ``How does knowledge graph embedding extrapolate to unseen data: a semantic evidence view,'' in \emph{Proceedings of the AAAI Conference on Artificial Intelligence}, vol.~36, no.~5, 2022, pp. 5781--5791.

\bibitem{song2021rot}
T.~Song, J.~Luo, and L.~Huang, ``Rot-pro: Modeling transitivity by projection in knowledge graph embedding,'' \emph{Advances in Neural Information Processing Systems}, vol.~34, pp. 24\,695--24\,706, 2021.

\bibitem{DBLP:journals/corr/YangYHGD14a}
\BIBentryALTinterwordspacing
B.~Yang, W.~Yih, X.~He, J.~Gao, and L.~Deng, ``Embedding entities and relations for learning and inference in knowledge bases,'' in \emph{3rd International Conference on Learning Representations, {ICLR} 2015, San Diego, CA, USA, May 7-9, 2015, Conference Track Proceedings}, Y.~Bengio and Y.~LeCun, Eds., 2015. [Online]. Available: \url{http://arxiv.org/abs/1412.6575}
\BIBentrySTDinterwordspacing

\bibitem{dettmers2018convolutional}
T.~Dettmers, P.~Minervini, P.~Stenetorp, and S.~Riedel, ``Convolutional 2d knowledge graph embeddings,'' in \emph{Proceedings of the AAAI conference on artificial intelligence}, vol.~32, no.~1, 2018.

\bibitem{toutanova-etal-2016-compositional}
\BIBentryALTinterwordspacing
K.~Toutanova, V.~Lin, W.-t. Yih, H.~Poon, and C.~Quirk, ``Compositional learning of embeddings for relation paths in knowledge base and text,'' in \emph{Proceedings of the 54th Annual Meeting of the Association for Computational Linguistics (Volume 1: Long Papers)}.\hskip 1em plus 0.5em minus 0.4em\relax Berlin, Germany: Association for Computational Linguistics, Aug. 2016, pp. 1434--1444. [Online]. Available: \url{https://aclanthology.org/P16-1136}
\BIBentrySTDinterwordspacing

\bibitem{das-etal-2017-chains}
\BIBentryALTinterwordspacing
R.~Das, A.~Neelakantan, D.~Belanger, and A.~McCallum, ``Chains of reasoning over entities, relations, and text using recurrent neural networks,'' in \emph{Proceedings of the 15th Conference of the {E}uropean Chapter of the Association for Computational Linguistics: Volume 1, Long Papers}.\hskip 1em plus 0.5em minus 0.4em\relax Valencia, Spain: Association for Computational Linguistics, Apr. 2017, pp. 132--141. [Online]. Available: \url{https://aclanthology.org/E17-1013}
\BIBentrySTDinterwordspacing

\bibitem{gu2015traversing}
K.~Guu, J.~Miller, and P.~Liang, ``Traversing knowledge graphs in vector space,'' in \emph{Empirical Methods in Natural Language Processing (EMNLP)}, 2015.

\bibitem{lao-etal-2011-random}
\BIBentryALTinterwordspacing
N.~Lao, T.~Mitchell, and W.~W. Cohen, ``Random walk inference and learning in a large scale knowledge base,'' in \emph{Proceedings of the 2011 Conference on Empirical Methods in Natural Language Processing}.\hskip 1em plus 0.5em minus 0.4em\relax Edinburgh, Scotland, UK.: Association for Computational Linguistics, Jul. 2011, pp. 529--539. [Online]. Available: \url{https://aclanthology.org/D11-1049}
\BIBentrySTDinterwordspacing

\bibitem{neelakantan-etal-2015-compositional}
\BIBentryALTinterwordspacing
A.~Neelakantan, B.~Roth, and A.~McCallum, ``Compositional vector space models for knowledge base completion,'' in \emph{Proceedings of the 53rd Annual Meeting of the Association for Computational Linguistics and the 7th International Joint Conference on Natural Language Processing (Volume 1: Long Papers)}.\hskip 1em plus 0.5em minus 0.4em\relax Beijing, China: Association for Computational Linguistics, Jul. 2015, pp. 156--166. [Online]. Available: \url{https://aclanthology.org/P15-1016}
\BIBentrySTDinterwordspacing

\bibitem{liu2020path}
W.~Liu, A.~Daruna, Z.~Kira, and S.~Chernova, ``Path ranking with attention to type hierarchies,'' in \emph{Proceedings of the AAAI Conference on Artificial Intelligence}, vol.~34, no.~03, 2020, pp. 2893--2900.

\bibitem{yang2017differentiable}
F.~Yang, Z.~Yang, and W.~W. Cohen, ``Differentiable learning of logical rules for knowledge base reasoning,'' \emph{Advances in neural information processing systems}, vol.~30, 2017.

\bibitem{rocktaschel2017end}
\BIBentryALTinterwordspacing
T.~Rockt{\"{a}}schel and S.~Riedel, ``End-to-end differentiable proving,'' in \emph{Advances in Neural Information Processing Systems 30: Annual Conference on Neural Information Processing Systems 2017, 4-9 December 2017, Long Beach, CA, {USA}}, 2017, pp. 3791--3803. [Online]. Available: \url{http://papers.nips.cc/paper/6969-end-to-end-differentiable-proving}
\BIBentrySTDinterwordspacing

\bibitem{fu-etal-2019-collaborative}
\BIBentryALTinterwordspacing
C.~Fu, T.~Chen, M.~Qu, W.~Jin, and X.~Ren, ``Collaborative policy learning for open knowledge graph reasoning,'' in \emph{Proceedings of the 2019 Conference on Empirical Methods in Natural Language Processing and the 9th International Joint Conference on Natural Language Processing (EMNLP-IJCNLP)}.\hskip 1em plus 0.5em minus 0.4em\relax Hong Kong, China: Association for Computational Linguistics, Nov. 2019, pp. 2672--2681. [Online]. Available: \url{https://aclanthology.org/D19-1269}
\BIBentrySTDinterwordspacing

\bibitem{florence2022implicit}
P.~Florence, C.~Lynch, A.~Zeng, O.~A. Ramirez, A.~Wahid, L.~Downs, A.~Wong, J.~Lee, I.~Mordatch, and J.~Tompson, ``Implicit behavioral cloning,'' in \emph{Conference on Robot Learning}.\hskip 1em plus 0.5em minus 0.4em\relax PMLR, 2022, pp. 158--168.

\bibitem{osa2018algorithmic}
T.~Osa, J.~Pajarinen, G.~Neumann, J.~A. Bagnell, P.~Abbeel, J.~Peters \emph{et~al.}, ``An algorithmic perspective on imitation learning,'' \emph{Foundations and Trends{\textregistered} in Robotics}, vol.~7, no. 1-2, pp. 1--179, 2018.

\bibitem{ho2016generative}
J.~Ho and S.~Ermon, ``Generative adversarial imitation learning,'' \emph{Advances in neural information processing systems}, vol.~29, 2016.

\bibitem{hochreiter1997long}
S.~Hochreiter and J.~Schmidhuber, ``Long short-term memory,'' \emph{Neural computation}, vol.~9, no.~8, pp. 1735--1780, 1997.

\bibitem{NIPS2013_1cecc7a7}
\BIBentryALTinterwordspacing
A.~Bordes, N.~Usunier, A.~Garcia-Duran, J.~Weston, and O.~Yakhnenko, ``Translating embeddings for modeling multi-relational data,'' in \emph{Advances in Neural Information Processing Systems}, C.~Burges, L.~Bottou, M.~Welling, Z.~Ghahramani, and K.~Weinberger, Eds., vol.~26.\hskip 1em plus 0.5em minus 0.4em\relax Curran Associates, Inc., 2013. [Online]. Available: \url{https://proceedings.neurips.cc/paper/2013/file/1cecc7a77928ca8133fa24680a88d2f9-Paper.pdf}
\BIBentrySTDinterwordspacing

\end{thebibliography}


\end{document}


\maketitle

\appendix
\begin{figure*}[h!] 
    \centering
    \includegraphics[scale=0.4]{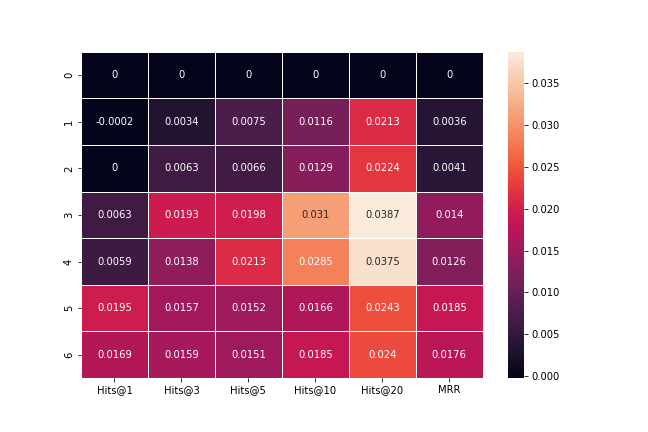}\hspace{-15mm}
    \includegraphics[scale=0.4]{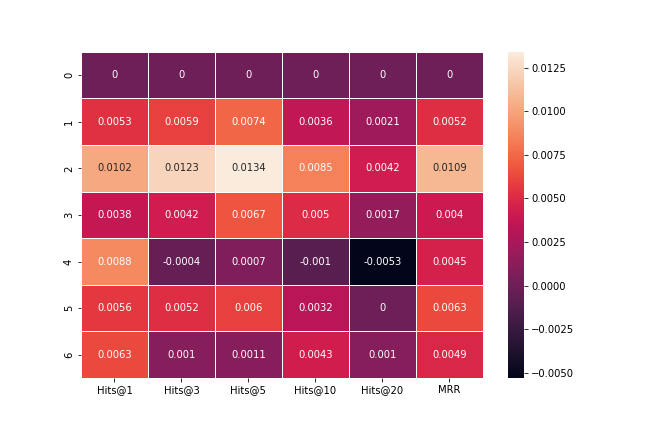}\hspace{-15mm}\\
    (a) FB15K-237  \hspace{50mm}  (b) WN18RR\\
    \includegraphics[scale=0.4]{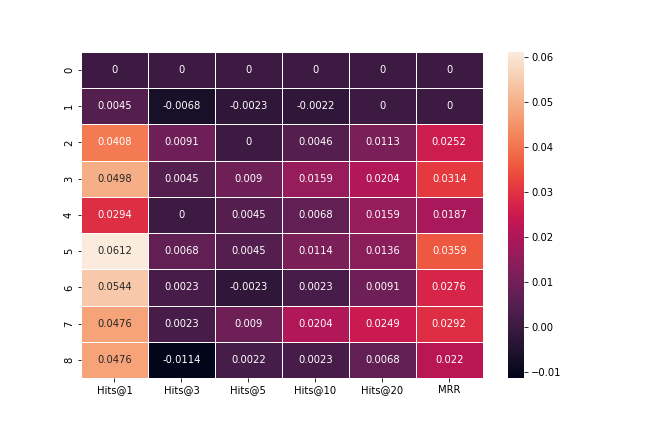}\hspace{-15mm}
    \includegraphics[scale=0.4]{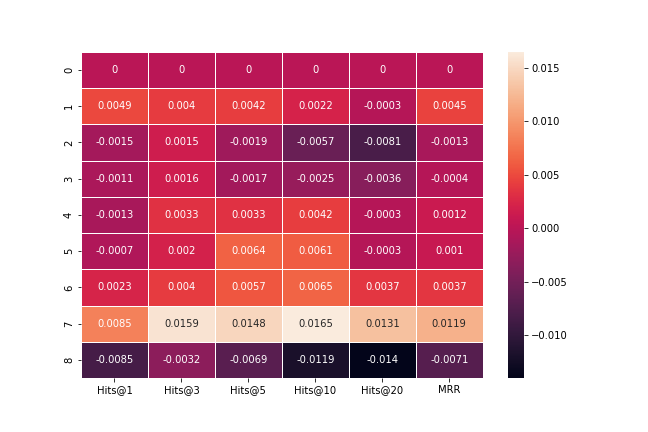}\\
    (c) NELL-995 \hspace{50mm} (d) FB60K
    \caption{Heatmap of Hits@k and MRR metrics for different SL training epochs followed by RL training. MINERVA is the baseline (SS-MINERVA) here. Row 0 is pure RL. The value in each cell is the score for that cell minus the score for that category at epoch 0 to highlight the difference.}
    \label{fig:heatmap_all}
\end{figure*}
\section{Appendix 1 Hyperparameter}

\paragraph{RL agent} The hyperparameters related to the RL part of our SS-MINERVA and SS-MultihopKG are the same as the original paper for FB15K-237, NELL-995 and WN18RR. 

Since this is the first work applying MINERVA to FB60K, we tried our best to adjust the hyperparameters to achieve the best performance. The size of the network is the same as FB15K-237. We also use the Wavier initialization for the embedding and neural network. The weight parameter $\beta$ for the entropy regularization term is 0.02 and the learning rate is $1^{-3}$.

\paragraph{SL agent} We keep all the hyperparamters for SL agent such as the the number of layers, the number of neurons in each layer, the paramters of LSTM the same as the RL agent. The learning rate for the SL agent is $1^{-3}$.

\section{Appendix 2 Performance analysis}
\label{sec:appendix}
In Fig. \ref{fig:heatmap_all}, we show the heatmaps of Hits@k and MRR metrics for all four datasets (we only show the heatmaps of FB15K-237 and FB60K in the main manuscript). From the results, we can see that the agent achieves its best performance with three, two, five, and seven epochs of SL pretraining on FB1K-237, WN18RR, NELL-995 and FB60K respectively. The results reported in Table 2 and 3 are based on these SL training epochs.

From Table 2 and Table 3 in the main manuscript, we can see that the improvements of SSRL to RL on different KG datsets are quite different. The improvements on FB15K-237 and NELL-995 datasets are significant. However, the improvement on WN18RR is mediocre and FB60K is moderate. We think there are two reasons for it: 
\begin{itemize}
    \item The imbalanced training data for different relation types.
    \item The small graph degree (number of edges connecting to each edge);
\end{itemize}

\begin{figure*}
    \centering
    \includegraphics[scale=0.6]{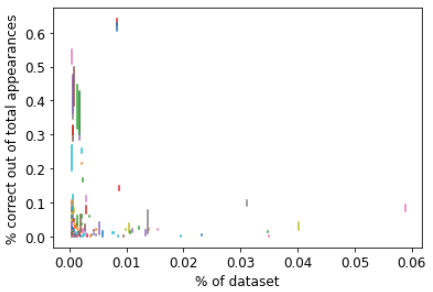}
    \includegraphics[scale=0.6]{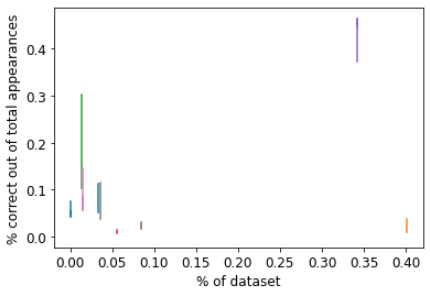}\\
    (a) FB15K-237  \hspace{50mm}  (b) WN18RR\\
    \includegraphics[scale=0.6]{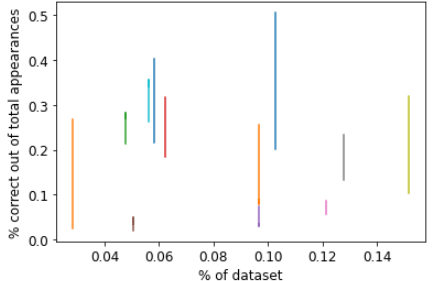}
    \includegraphics[scale=0.6]{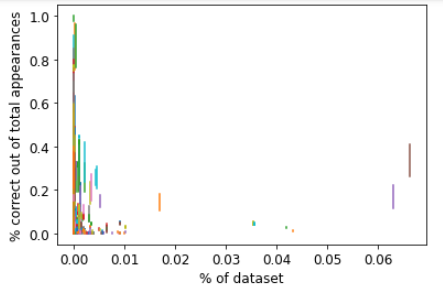}\\
    (c) NELL-995 \hspace{50mm} (d) FB60K
    \caption{Distribution of relations in the training set. The vertical lines correspond to the accuracy of the agent on that relation increasing over the course of training}
    \label{fig:relation distribtion}
\end{figure*}

In order to verify this two assumptions, we show the distribution of relations in the training set in Fig.\ref{fig:relation distribtion}. On NELL, which the agent performs the best on out of all the datasets, there are an even distribution of relation types that show up with different frequencies in the dataset. This means that the agent is seeing a good sample of the different query types, and trains well on all of them. On WN18RR and FB60K, the distributions of relation types are very skewed. 

For WN18RR, where the improvement of SSRL on RL is most insignificant, just two relations types comprise more than 75\% of the dataset while the others show up comparatively infrequently. Since the goal of SL is to learn the underlying distribution based on labeled data and in doing encourage the agent to explore paths it wouldn't otherwise, pretraining on these skewed distributions would encourage the agent to find more correct paths on the dominate relations, so it achieves higher overall accuracy on the whole training set. However, it may sacrifice the performance on less common relations where finding even one correct path is difficult. Furthermore, the idea behind SL is that by teaching the agent this underlying distribution it may take potentially advantageous paths it otherwise wouldn't take; if the the dataset is nostly homogenous (e.g. sharply skewed towards a select few relations), there is little that SL could show the agent that it wouldn't find during RL anywways. As the a result, SL pretraining does not help in terms of Hits@k when k is large on sharply skewed datasets like WN18RR 

For FB60K, Fig.\ref{fig:heatmap_all} shows that the trial where a correct path was found for the highest percentage of queries uses 7000 steps of SL. The graphs in Fig.\ref{fig:learing curve} in the main manuscript track average reward over the batch, which includes multiple trials for each query. Our analysis of agent performance by relation type included in Fig.\ref{fig:relation distribtion} shows that FB60K is the only dataset containing numerous relations with a near 100\% fact prediction success rate, implying that it contains facts easier for the agent to understand. Intuitively, the immediately strong performance on these "low hanging fruit" will increase the average performance towards the beginning of training, but as increasing durations of SL pretraining force the agent to generalize to more complex relations its success rate on these relations naturally falls, resulting in a decreasing average performance and an increasing percentage of queries for which a correct path is found. This tracks with the graph and heatmap presented in Fig.\ref{fig:learing curve} in the main manuscript and Fig.\ref{fig:heatmap_all}.

\begin{figure*}[h] 
    \centering
    \includegraphics[scale=0.14]{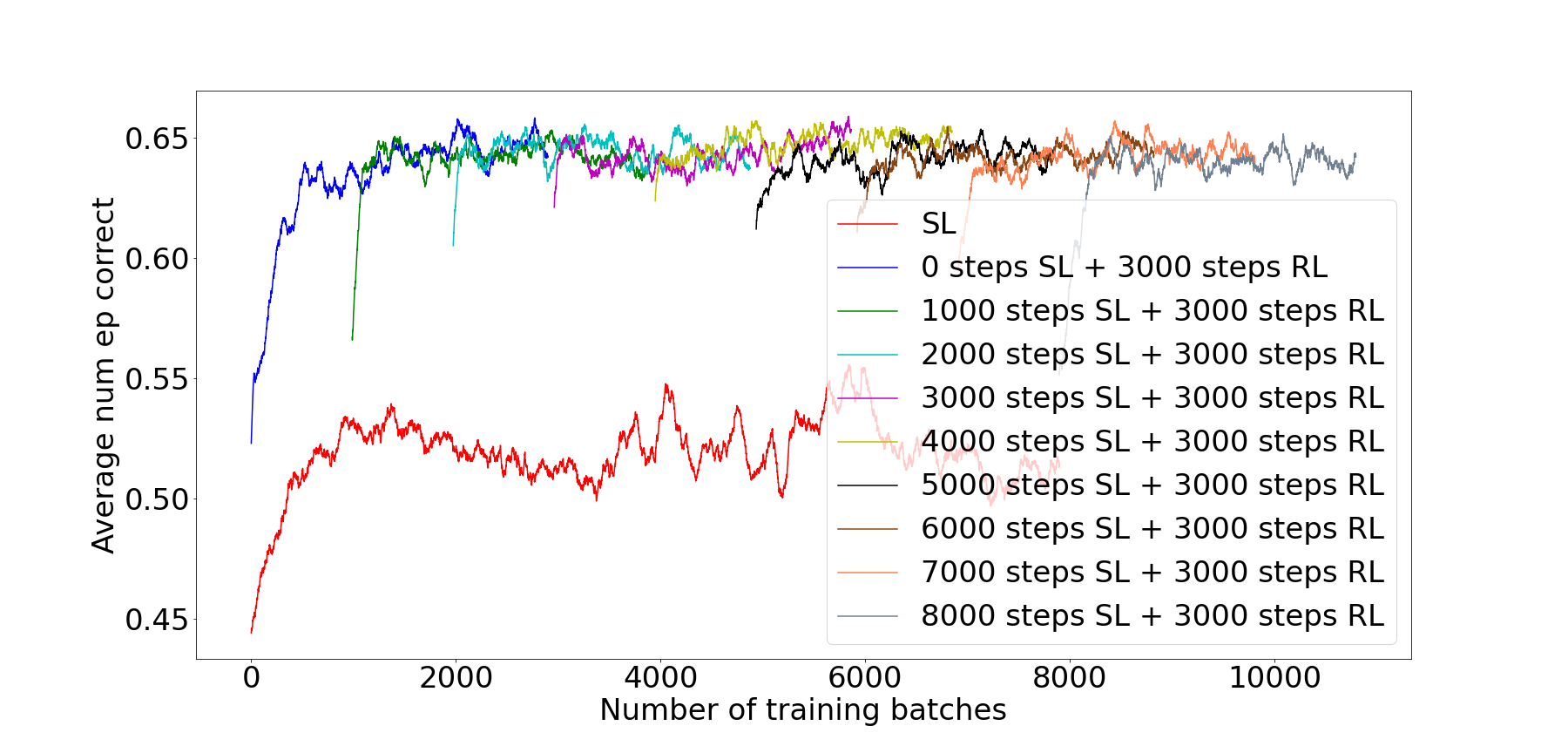}\hspace{-10mm} 
    \includegraphics[scale=0.14]{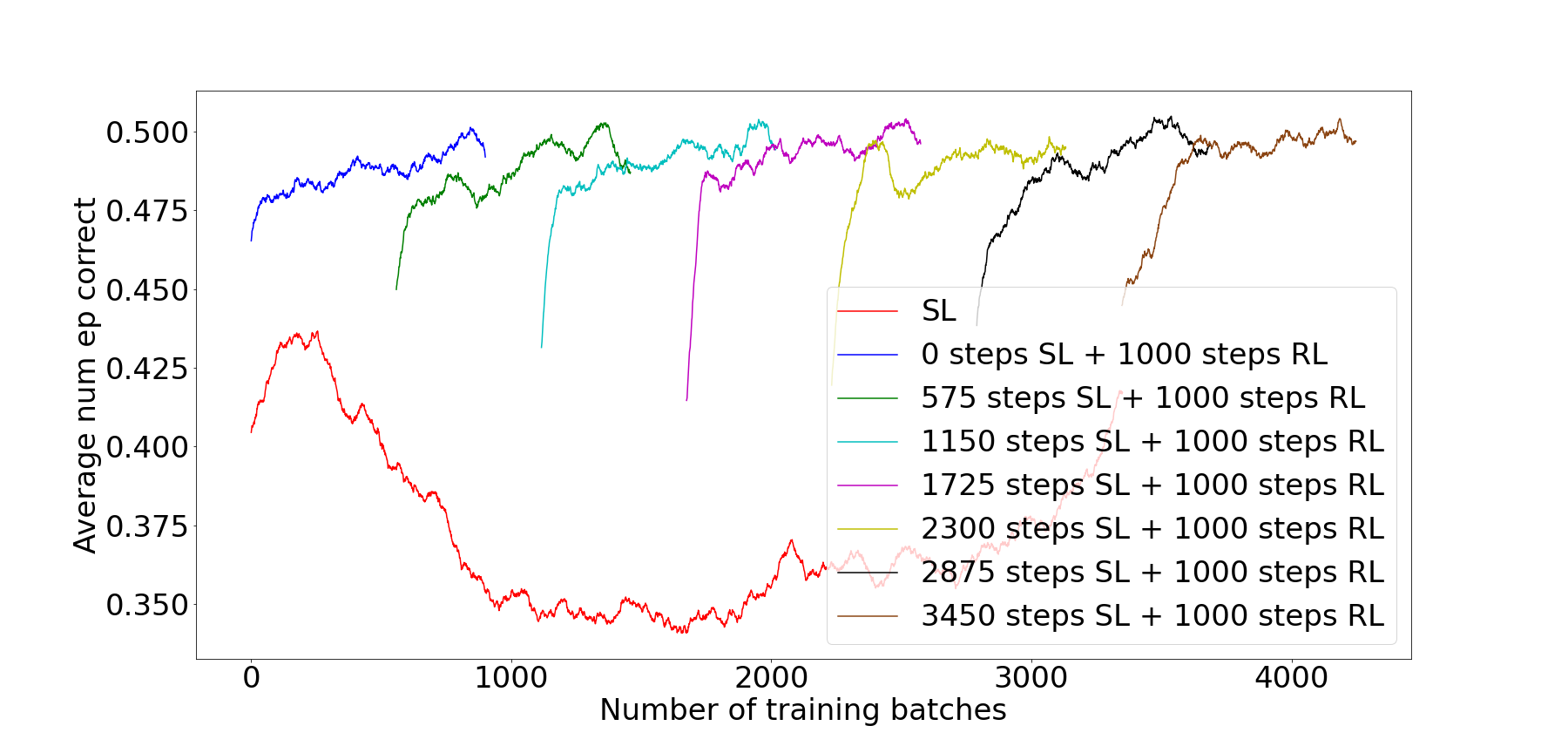}\\
    (a) FB15K-237  \hspace{45mm}  (b) WN18RR\\
    \includegraphics[scale=0.14]{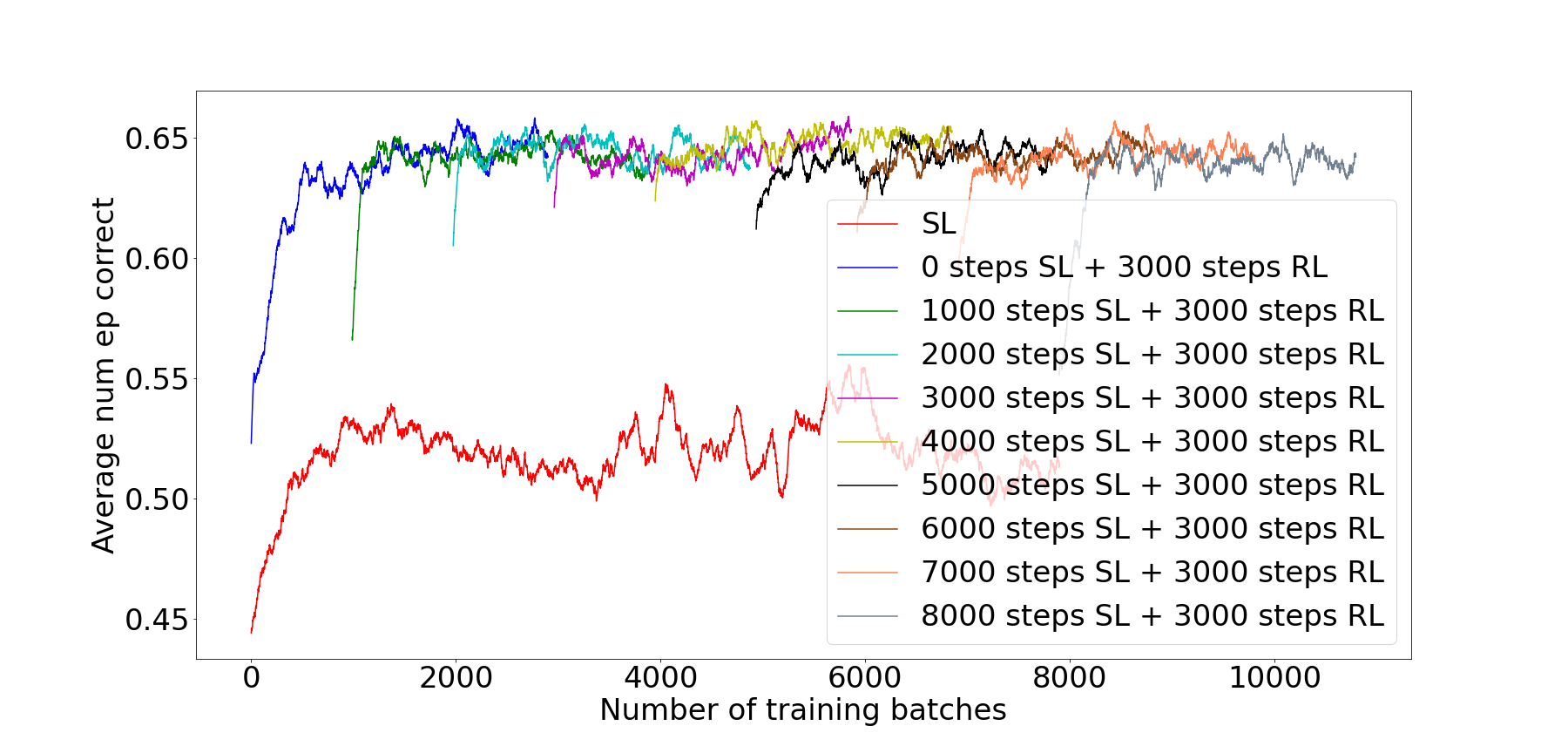}\hspace{-10mm} 
    \includegraphics[scale=0.14]{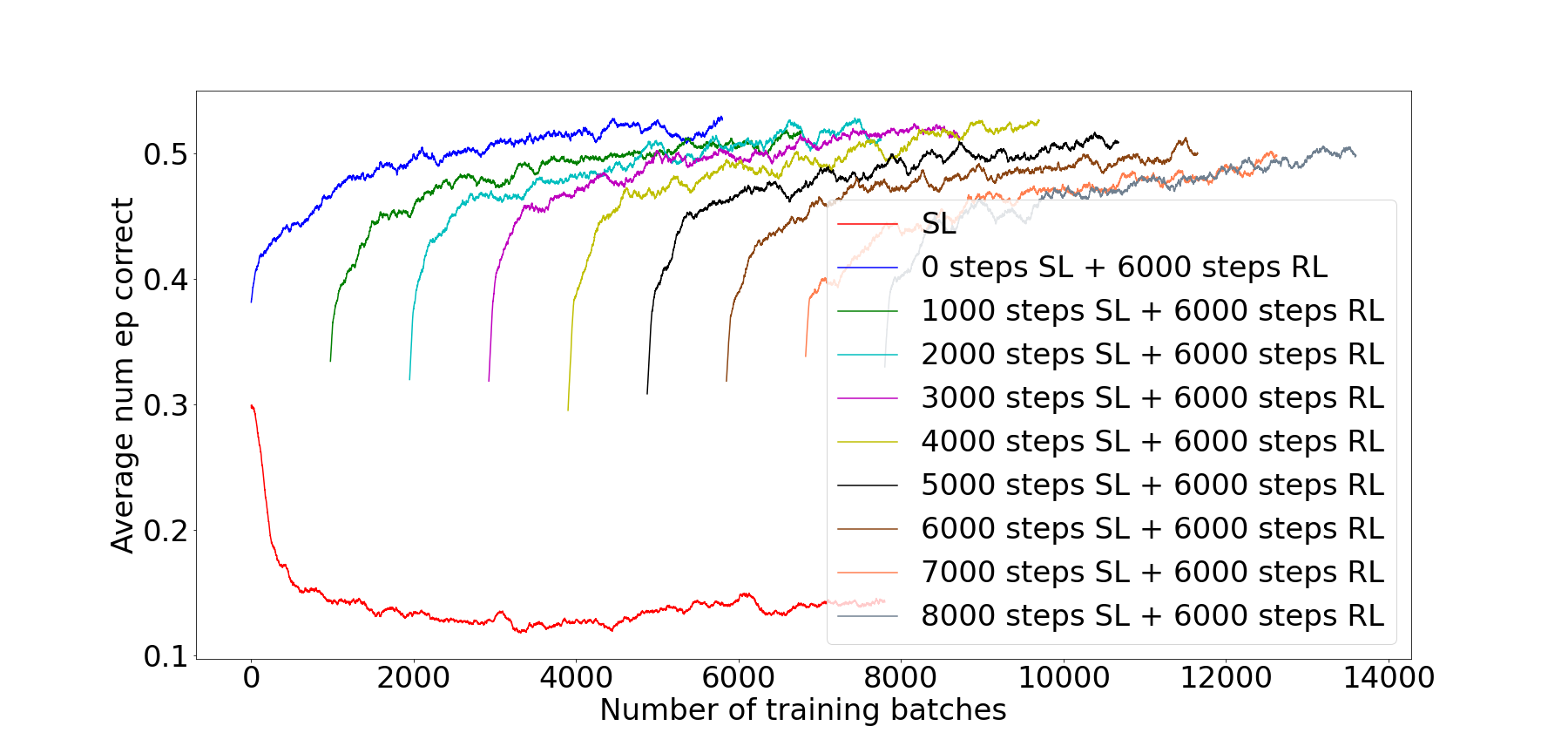}\\
    (c) NELL-995 \hspace{45mm} (d) FB60K
    \caption{Learning curves ((Hits@20 v.s. number of training batches)) with different SL pretraining steps followed by same RL steps}
    \label{fig:learing curve2}
\end{figure*}

To backup these claim, we plot the learning curves with Hit@20 as the metric in Fig.\ref{fig:learing curve2}. The SL graphs show negative trends on skewed datasets, i.e., WN18RR and FB60K. This is reasonable because the target of SL is to minimize the crossentropy loss, therefore maximize the overall accuracy. It is noted that the final performance on Hits@20 can be fixed by the RL training stage. This also shows to importance of the RL training stage in the proposed architecture.

We also observed that our SSRL architecture performs very well on FB15K-237 where the relation type distribution is also skewed (although it is less skewed than FB60K and WN18RR). We believe the excellent performance is due to the high graph degree where the SL are particularly helpful. Table 1 of the main manuscript shows the median degree of FB15K-237 is 14 whereas other datasets is less than 4.

In summary, the extent of improvement of SSRL to RL depends on the statistics of the datasets such as graph degree and relation distributions. Currently, our SSRL architecture generates labels for random relation types. In our future work, we will discuss how to generate labels based on the relation disctributions in each dataset therefore improve the performance further.

